\pdfoutput=1
\documentclass[11pt]{article}

\usepackage[final]{acl}

\usepackage{times}
\usepackage{latexsym}
\usepackage[T1]{fontenc}
\usepackage[utf8]{inputenc}

\usepackage[]{natbib}
\usepackage{hyperref}

\usepackage{microtype}

\usepackage{inconsolata}

\usepackage{graphicx}

\usepackage{booktabs}
\usepackage{amsmath, amssymb}
\usepackage{enumitem}
\usepackage{url}
\usepackage{comment}
\usepackage[table,dvipsnames]{xcolor}

\usepackage{longtable}
\usepackage{booktabs}  % For professional-looking tables
\usepackage{array}     % For more column formatting options
\usepackage{ragged2e}

\usepackage[table,dvipsnames]{xcolor}
\usepackage{fvextra}
\usepackage{xspace} 
\usepackage{multirow}
\usepackage{ulem}
\usepackage{fvextra}  % loads fancyvrb
\usepackage[most]{tcolorbox}
\usepackage{tabto}
\TabPositions{1.5em}
\usepackage{xcolor}
\usepackage{xspace}

\usepackage{booktabs}
\usepackage{tabularx}
\usepackage{ragged2e}
\usepackage{ltablex}
\usepackage{array}
\keepXColumns
\newcolumntype{P}[1]{>{\RaggedRight\arraybackslash}p{#1}}

\usepackage{booktabs}
\usepackage{ragged2e}
\usepackage{ltablex}
\usepackage{array}
\keepXColumns

\usepackage{graphicx}     % For \includegraphics
\usepackage{subcaption}   % For subfigures
\usepackage{caption}      % Recommended with subcaption

\definecolor{lightblue}{RGB}{235, 245, 255}
\newtcolorbox{promptbox}{
  colback=lightblue,
  colframe=RoyalBlue!75!white,
  fonttitle=\bfseries,
  title=System Prompt,
  boxsep=5pt,
  arc=4pt,
  boxrule=1pt,
  breakable
}

\newcommand{\ssfTaxonomy}{\textsc{SSF-Taxonomy}\xspace}

\newcommand{\ssfFormalisms}{\textsc{SocialStoryFrames}\xspace}
\newcommand{\ssfGenerator}{\textsc{SSF-Generator}\xspace}
\newcommand{\ssfClassifier}{\textsc{SSF-Classifier}\xspace}
\newcommand{\ssfCorpus}{\textsc{SSF-Corpus}\xspace}
\newcommand{\ssfSplitCorpus}{\textsc{SSF-Split-Corpus}\xspace}
\newcommand{\ssfStratifiedCorpus}{\textsc{SSF-Stratified-Corpus}\xspace}
\newcommand{\refClassifier}{\textsc{GPT-4.1 (K-Shot)}\xspace}
\newcommand{\ssfCorpusSize}{$6,140$\xspace}
\newcommand{\ssfSplitCorpusSize}{$1,778$\xspace}
\newcommand{\ssfStratifiedCorpusSize}{$2,250$\xspace}
\newcommand{\ssfSimMetric}{\texttt{ssf-sim}\xspace}

\definecolor{joel-color}{HTML}{7881F2}

\definecolor{goal-color}{HTML}{f3d0ee}
\definecolor{intent-color}{HTML}{c7eefb}
\definecolor{emotion-color}{HTML}{c3f1c8}
\definecolor{character-color}{HTML}{bcf0e6}
\definecolor{moral-color}{HTML}{b3cfed}
\definecolor{stance-color}{HTML}{cbe9a0}
\definecolor{feeling-color}{HTML}{d0cbf1}
\definecolor{aesthetic-color}{HTML}{f0bfc3}
\definecolor{causal-color}{HTML}{efe6ad}
\definecolor{prediction-color}{HTML}{fbe2d6}

\usepackage{booktabs}
\usepackage{pgf}
\usepackage{colortbl}

\definecolor{color1}{HTML}{93003a}
\definecolor{color2}{HTML}{cf3759}
\definecolor{color3}{HTML}{f4777f}
\definecolor{color4}{HTML}{ffbcaf}
\definecolor{color5}{HTML}{ffffe0}
\definecolor{color6}{HTML}{a5d5d8}
\definecolor{color7}{HTML}{73a2c6}
\definecolor{color8}{HTML}{4771b2}
\definecolor{color9}{HTML}{00429d}

\usepackage{array}

\newcommand*{\opacity}{30}% Here you can change the opacity of the background color!
\newcommand*{\minval}{0.50}% Define the minimum value on your data set
\newcommand*{\maxval}{1.0}% Define the maximum value in your data set!

\newcommand{\gradient}[1]{%
    \ifdimcomp{#1pt}{>}{\maxval pt}{#1}{%
        \ifdimcomp{#1pt}{<}{\minval pt}{#1}{%
            \pgfmathparse{int(round(8*(#1/(\maxval-\minval))-(\minval*(8/(\maxval-\minval)))))}%
            \xdef\tempa{\pgfmathresult}%
            \ifcase\tempa
                \colorbox{color1!\opacity}{#1}\or
                \colorbox{color2!\opacity}{#1}\or
                \colorbox{color3!\opacity}{#1}\or
                \colorbox{color4!\opacity}{#1}\or
                \colorbox{color5!\opacity}{#1}\or
                \colorbox{color6!\opacity}{#1}\or
                \colorbox{color7!\opacity}{#1}\or
                \colorbox{color8!\opacity}{#1}\or
                \colorbox{color9!\opacity}{#1}%
            \fi
    }}%
}

\newcommand*{\minvall}{-1.0}% Define the minimum value on your data set
\newcommand*{\maxvall}{1.0}% Define the maximum value in your data set!

\newcommand{\gradientt}[1]{%
    \ifdimcomp{#1pt}{>}{\maxvall pt}{#1}{%
        \ifdimcomp{#1pt}{<}{\minvall pt}{#1}{%
            \pgfmathparse{int(round(8*(#1/(\maxvall-\minvall))-(\minvall*(8/(\maxvall-\minvall)))))}%
            \xdef\tempa{\pgfmathresult}%
            \ifcase\tempa
                \colorbox{color1!\opacity}{#1}\or
                \colorbox{color2!\opacity}{#1}\or
                \colorbox{color3!\opacity}{#1}\or
                \colorbox{color4!\opacity}{#1}\or
                \colorbox{color5!\opacity}{#1}\or
                \colorbox{color6!\opacity}{#1}\or
                \colorbox{color7!\opacity}{#1}\or
                \colorbox{color8!\opacity}{#1}\or
                \colorbox{color9!\opacity}{#1}%
            \fi
    }}%
}

\title{Social Story Frames:\\Contextual Reasoning about Narrative Intent and Reception}

\newcommand{\aspace}{\hspace{1em}}
\newcommand{\cmu}{$^\clubsuit$}
\newcommand{\cu}{$^\spadesuit$}
\newcommand{\umf}{$^\diamondsuit$}
\newcommand{\uc}{$^\heartsuit$}
\newcommand{\mcgill}{$^\triangle$}

\author{
Joel Mire\cmu \aspace 
Maria Antoniak\cu \aspace 
Steven R. Wilson\umf \aspace 
Zexin Ma\uc \aspace
\vspace{.3em} \\
\textbf{Achyutarama R. Ganti}\umf \aspace
\textbf{Andrew Piper}\mcgill \aspace 
\textbf{Maarten Sap}\cmu
\vspace{.3em}\\
\small
\cmu Carnegie Mellon University 
\aspace \cu University of Colorado Boulder
\aspace \umf University of Michigan-Flint \\
\small
\aspace \uc University of Connecticut
\aspace \mcgill McGill University
}

\DefineVerbatimEnvironment{wrapverbatim}{Verbatim}
  {breaklines=true, breakanywhere=false}
  
\begin{document}

\maketitle

\begin{abstract}
Reading stories evokes rich interpretive, affective, and evaluative responses, such as inferences about narrative intent or judgments about characters. Yet, computational models of reader response are limited, preventing nuanced analyses. To address this gap, we introduce \mbox{\ssfFormalisms}, a formalism for distilling plausible inferences about reader response, such as perceived author intent, explanatory and predictive reasoning, affective responses, and value judgments, using conversational context and a taxonomy grounded in narrative theory, linguistic pragmatics, and psychology. We develop two models, \ssfGenerator and \ssfClassifier, validated through human surveys ($N=382$ participants) and expert annotations, respectively. We conduct pilot analyses to showcase the utility of the formalism for studying storytelling at scale. Specifically, applying our models to \ssfCorpus, a curated dataset of \ssfCorpusSize social media stories from diverse contexts, we characterize the frequency and interdependence of storytelling intents, and we compare and contrast narrative practices (and their diversity) across communities. By linking fine-grained, context-sensitive modeling with a generic taxonomy of reader responses, \ssfFormalisms enable new research into storytelling in online communities.
\end{abstract}

\section{Introduction}
When reading stories, people draw rich inferences and evaluations \cite{Graesser1994-gz, Goldman2015-br} and have varied affective responses \cite{Miall2002-cj, Hamby2023-ww} shaped by both the text and contextual factors that extend beyond the text itself \cite{Prince1983-qv}.

\begin{figure}[t]
    \centering
    \includegraphics[width=\columnwidth]{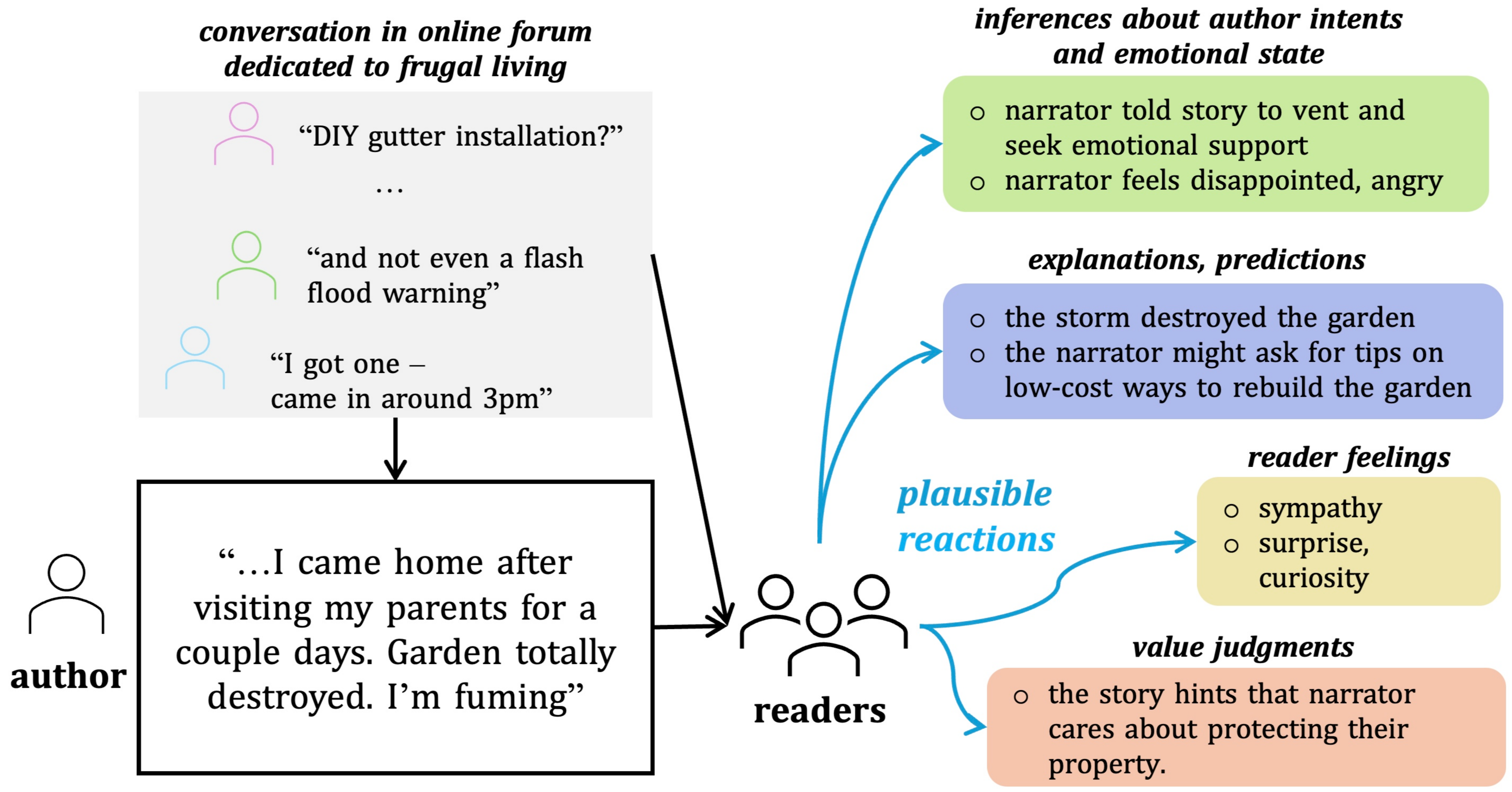}
    \caption{Storytelling evokes many forms of reader response. We introduce \ssfFormalisms, a formalism for reasoning about intent and reception of conversational stories on social media.}
\label{fig:example-overview}
\end{figure}

For example, suppose someone tells a story about how they arrived home to find their garden destroyed in an online forum dedicated to budget-friendly home projects. Many readers might sympathetically perceive the author as disappointed, angry, and seeking emotional support, while also inferring why the garden was destroyed and predicting that the author might subsequently ask about low-cost ways to rebuild the garden (Fig. \ref{fig:example-overview}). %\looseness=-1
Studying these extra-textual responses at scale can deepen understanding of the social dynamics of online storytelling, which plays a central role in how people communicate, relate to others, and make sense of experiences \cite{Bietti2019-vp}.

Existing methods, however, are limited by a tradeoff between depth and breadth. Studies of individual communities offer rich insights---for example, how branching narratives reflect negotiations of power in birth stories \citep{Antoniak2019-ll}, or how COVID-19 quarantine stories grapple with uncertainty \cite{Ho2024-jd}---but these methods are difficult to generalize. In contrast, large-scale analyses often rely on coarse measures, such as storytelling frequency \citep{Antoniak2024-ko} or differences in structural prototypes \citep{Yan2019-zy}, sacrificing interpretive depth. We need scalable approaches for characterizing the social dynamics of storytelling across diverse online communities.

\begin{figure*}[t]
    \centering
    \includegraphics[width=\textwidth]{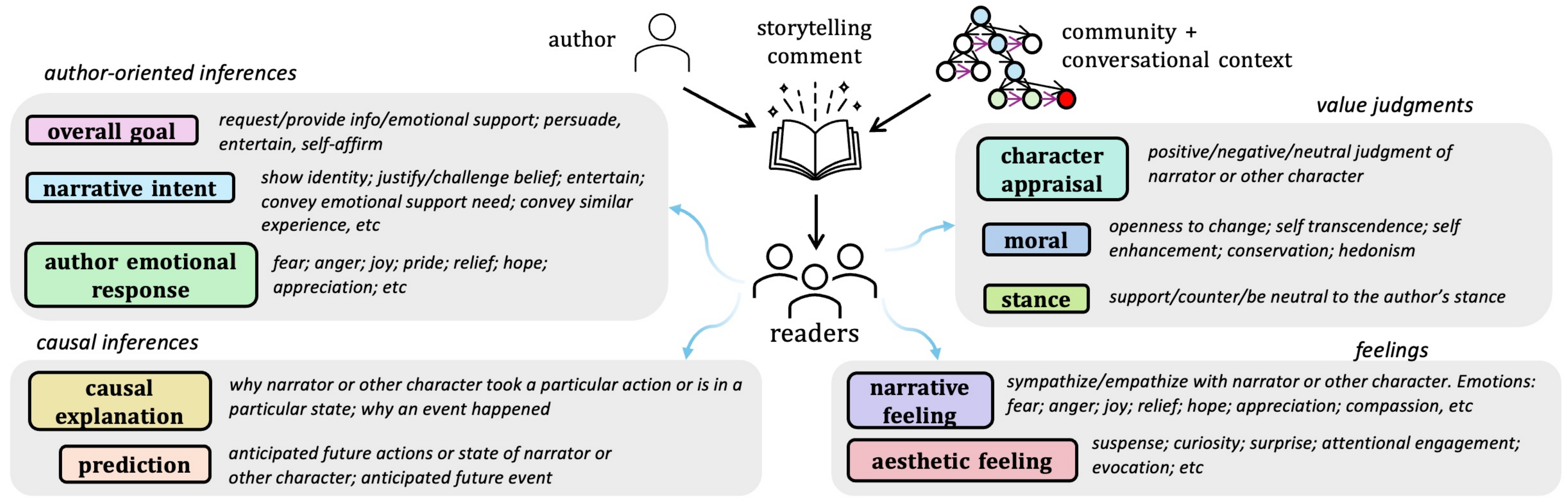}
    \caption{We introduce \ssfTaxonomy, a taxonomy of reader response to informal storytelling on social media.}
    \label{fig:taxonomy-overview}
\end{figure*}

Thus, we introduce \ssfFormalisms, a formalism for reasoning about intent and reception of conversational stories on social media, based on theories of narrative and reader response,  pragmatics, and psychology. Core to our approach is attention to the broader pragmatic frame in which stories are embedded, specifically, the community and conversational contexts in which readers encounter stories. Our formalism characterizes reader response in terms of our taxonomy, \ssfTaxonomy (Fig. \ref{fig:taxonomy-overview}), which encompasses $10$ dimensions of reader response, including author-oriented inferences (e.g., intent), explanatory/predictive inferences, affective responses, and value judgments. 

We operationalize \ssfFormalisms in a two-stage modeling pipeline encompassing inference generation and classification. Using supervised finetuning (SFT) based distillation of GPT-4o, we train a model, \ssfGenerator, to generate contextual inferences about reader response. Following a similar paradigm with GPT-4.1, we train a classifier, \ssfClassifier, to map inferences onto fine-grained \ssfTaxonomy subcategories. We validate inferences through human surveys for \ssfClassifier's training data ($N{=}278$ participants) and model outputs ($N{=}104$); inference classification is validated through expert annotation. 

To showcase the utility of \ssfFormalisms, we apply our modeling pipeline to construct \ssfCorpus, a corpus of \ssfCorpusSize stories and their contexts across a multi-community social media platform. We characterize the basic social functions of narrative---as evidence, entertainment, and means for self-expression and connection---and show how broad communicative goals (e.g., providing emotional support) are associated with particular narrative intents (e.g., conveying a similar experience). Additionally, we conduct a preliminary analysis using \ssfFormalisms to compare online communities based on their narrative practices. Our findings show that topically distinct communities (e.g., \texttt{r/MakeupAddiction} and \texttt{r/buildapc}) can exhibit similar social patterns of storytelling that are not captured by semantic similarity alone, and that communities vary widely in their motivations for and responses to stories. Through the lens of narrative intent and reception, we offer scalable, nuanced analyses of the communicative, interpretive, and social functions of storytelling across diverse online communities.\footnote{Code: \href{https://github.com/joel-mire/social-story-frames}{social-story-frames}. Data: \href{https://huggingface.co/datasets/joelmire/ssf-corpus}{SSF-Corpus}. Models: \href{https://huggingface.co/joelmire/llama3.1-8b-it-ssf-generator}{SSF-Generator}, \href{https://huggingface.co/joelmire/llama3.1-8b-it-ssf-classifier}{SSF-Classifier}.}

\section{Related Work}
We situate our work within computational approaches to narrative reception, then review contemporary commonsense reasoning research that informs our tasks and models.

\subsection{Reader Response Approaches in NLP}
Our work models inferences readers might make in response to everyday stories. Prior studies have modeled specific aspects of reader response, such as suspense, curiosity, and surprise \cite{Steg2022-ic}, or assessments of literary quality \cite{Moreira2023-iy}. \citet{De-Fina2016-ot} examines online story comment threads, characterizing readers' relative engagement with the story content (\textit{taleworld}) compared to the context of its telling (\textit{storyrealm}). 

While several approaches exist for studying collective forms of reader response in online communities, they are tailored for communities organized around a narratively inflected topic, like the COVID-19 origin story \citep{Shahsavari2020-ii}, or a single focal narrative text, like a book in review forums \citep{Holur2021-ib} or a source text in fanfiction communities \citep{Vadde2024-px}. These works have primarily focused on representations or perceptions of characters and/or narrative schemas, leaving many dimensions of reader response underexplored. Many questions remain around how to quantify variation across communities.

We address these gaps by introducing a broad, $10$-dimensional reader response taxonomy, validating its generalizability across diverse communities, and demonstrating its value for comparative analyses of narrative practices. 

\subsection{Commonsense Reasoning in NLP}
Our work builds on and extends various works on commonsense reasoning in NLP to enable rich contextual inference about narrative reception. 

Much prior work has revolved around creating large commonsense knowledge bases and encoding this knowledge in models \cite{Rashkin2018-vs, Sap2019-yg, Bosselut2019-xh, west-etal-2022-symbolic}, but these approaches have been based on short, decontextualized event descriptions.

Specialized knowledge bases and commonsense inference methods for narrative texts have focused on particular aspects of story understanding, such as causal explanation \cite{Mostafazadeh2016-ur, mostafazadeh-etal-2020-glucose, Bhagavatula2019-ae}, plot consistency \cite{Ahuja2025-vt}, and character psychology \cite{rashkin-etal-2018-modeling, Vijayaraghavan2021-ot}, yet leave several aspects of narrative reception unexplored (e.g., perceived narrative intent, aesthetic feelings), and do not account for narrative or social contexts outside the story content, which are essential in a social media context. Recent works have started incorporating social context (e.g., social situation, speaker background) into task-specific reasoning, e.g., social acceptability \cite{Pyatkin2022-gb, Rao2023-ya}. Notably, the CobraFrames formalism introduced a novel taxonomy and model for contextual inference about social biases and toxicity \cite{zhou-etal-2023-cobra}. Our work builds on and extends these efforts toward contextual reasoning about a distinct social phenomenon, namely, narratives.

\section{SSF Formalism: Taxonomy and Tasks}
In this section, we introduce the \ssfFormalisms formalism, including its taxonomy of reader response dimensions (\ssfTaxonomy) and inference generation and classification tasks.

\subsection{Taxonomy Dimensions}
\label{sec:taxonomy}
Drawing on narrative theory, discourse processing, and psychology, we taxonomize dimensions of perceived intent and reception for social media storytelling (Fig. \ref{fig:taxonomy-overview}). We include 10 dimensions of reader response, encompassing inferences about authors, explanatory/predictive inferences about content, value judgments of characters or themes, and affective responses. In our work, ``author'' refers to a person who makes a storytelling post or comment, and ``reader'' refers to a member of an online community who encounters that post or comment in its conversational context. We further refine each dimension into subdimensions based on a multidisciplinary literature review and consideration of our target domain (short, informal social-media stories). We outline each dimension and its key subdimensions below; see App. \ref{app:taxonomy} for the full taxonomy and App. \ref{app:background-social-functions} for background on prior efforts to categorize the functions of storytelling.

\paragraph{\colorbox{goal-color}{Overall goal}} is the communicative intent of a comment, grounded in the speech act theoretic view that language use signals intentional social action \cite{Austin1975-bt}. For sub-dimensions, we draw on prior work on motivational factors for posting on social media, including providing or seeking information, emotional support, or stories \cite{biyani-etal-2014-identifying, Purohit2015-kz, Yang2019-we}. We also include persuasion \cite{Tan2016-kn}, entertainment \cite{Moore2017-ue}, and self-expression \cite{Orehek2017-yh}.

\paragraph{\colorbox{intent-color}{Narrative intent}} is related to---but distinct from---the overall goal, in part because storytelling may only partially span a post or comment. While theorists have posited high-level social functions of storytelling \cite{BoydUnknown-fm, Felski2014-xd, Dillon-SUnknown-qx, Crowley2003-vf} and the cognitive and affective impacts of reading fiction \cite{Dodell-Feder2018-fr}, localized intents of everyday social-media stories are understudied. Our subdimensions reflect narrative uses for identity \cite{Bruner1991-ny, Somers1994-xm, Dillon-SUnknown-qx}, sense-making \cite{Bietti2019-vp, Antoniak2019-ll}, argumentation \cite{Green2000-nv, Braddock2016-el, Leslie2015-iw, Krause2020-kp, Polletta2011-kp}, entertainment \cite{Brewer1982-pm}, expressing intense emotions, revising understanding, demonstrating need for emotional support, and conveying similar experiences.

\paragraph{\colorbox{emotion-color}{Author emotional response}} refers to an emotional state attributed to the act of telling the story, motivated by work on psycho-therapeutic effects of expressive writing \cite{Pennebaker1999-tr}. For a discrete taxonomy of emotions, we use \citeposs{Nabi2002-ff} categories of negative (e.g., fear, anger, sadness) and positive (e.g., joy, pride, hope) emotions, while also adding subcategories for ``appreciation'' and ``connection''.

\paragraph{\colorbox{causal-color}{Causal explanation}} is an inference about why an event or state occurred (e.g., mental states explaining actions). Explanatory inference types are widely acknowledged across discourse processing theories, such as \citeposs{Graesser1994-gz} constructionist theory of narrative text comprehension. 

\paragraph{\colorbox{prediction-color}{Prediction}} infers what might happen next in or beyond the story. This dimension is consistent with predictive inference types in \citeposs{Graesser1994-gz} taxonomy (e.g., ``character emotional reaction''). We distinguish prediction categories by target (narrator, other character, etc.) and by whether the prediction pertains to a future state or event.

\paragraph{\colorbox{character-color}{Character appraisal}} involves normative judgments of characters \cite{Graesser1994-gz}, categorized as positive, neutral, or negative.

\paragraph{\colorbox{moral-color}{Moral}} represents thematic or moralizing inferences, which relate closely to values. For discrete analyses, we adopt the most abstract categories from \citeposs{Schwartz2012-xk} theory: self-enhancement, self-transcendence, conservation, hedonism, and openness to change.

\paragraph{\colorbox{stance-color}{Stance}} captures the degree to which readers support, oppose, or are neutral toward any opinion or perspective implied in a post or comment \cite{Mohammad2017-ss}.

\paragraph{\colorbox{feeling-color}{Narrative feeling}} concerns readers’ emotional engagement with the story \textit{content} \cite{Miall2002-cj}. Beyond character appraisal, narrative feelings are typically oriented ``toward specific aspects of the fictional event sequence'', e.g., in the form of sympathy or empathy toward a character. To operationalize this dimension, we combine the high-level feeling types identified by \citet{Miall2002-cj} with the discrete emotion categories proposed by \citet{Nabi2002-ff}.

\paragraph{\colorbox{aesthetic-color}{Aesthetic feeling}} describes responses to narrative style, or how a story is told \cite{Miall2002-cj}. We focus on prominent aesthetic responses studied in narrative theory and empirical literary studies. These include suspense, curiosity, and surprise \cite{sternberg_how_2001}, and aspects of narrative absorption, including attention, transportation, and evocation \cite{Kuijpers2021-os}.\footnote{A final aspect of narrative absorption from \citet{Kuijpers2021-os}, emotional engagement, is largely covered by the narrative feeling dimension.}

\subsection{Tasks}
Based on our taxonomy, we define two tasks that encompass \ssfFormalisms reasoning.

\paragraph{Inference Generation} 
Inference generation produces a plausible free-text statement predicting how readers might respond to an author's story along a specific taxonomy dimension. The task is as follows: given a storytelling comment, its community and conversational context, and a brief description of the target dimension, generate a templated inference that reflects a likely reader response.

\paragraph{Inference Classification}
The second task maps the templated inference generations onto the fine-grained taxonomy, explicitly representing narrative reception in terms of our categorical theoretical constructs. This complements the information-dense free-text inferences, which are more grounded in the semantics of the story and its conversational context. We define the task as multi-label classification: given a dimension-specific inference, output zero or more subdimension labels.

\section{Stories Data}
We construct \ssfCorpus, a set of social media stories in context, on which we instantiate \ssfFormalisms.
We sample the data from a curated subset of the \textsc{reddit-corpus-small} dataset from ConvoKit~\citep{chang2020convokit}.

We preprocess and filter this dataset's 10,000 Reddit threads drawn from 100 subreddits by masking PII, including only texts containing stories, according to the \texttt{storyseeker} classifier \citep{Antoniak2024-ko}, and at least 175 characters, removing texts predicted to be toxic or sexually explicit, and excluding irrelevant subreddits. A full description of the curation process is in App. \ref{appendix-dataset-curation}, and we formally define the dataset structure in App. \ref{appendix-dataset-structure}.

The final corpus comprises \ssfCorpusSize stories along with their preceding context posts and comments. We define two variants of this corpus. The first, \ssfSplitCorpus (N=\ssfSplitCorpusSize), is partitioned into train, validation, and test splits,\footnote{$2/3$ train, $1/6$ validation, $1/6$ test split} stratified across subreddits with at least 45 stories. To ensure that our evaluation accounts not only for unseen stories in seen communities but also for unseen stories in unseen communities, $~10\%$ of the stories in each of the validation and test splits are drawn from $5$ subreddits that are absent from both the training split and the other evaluation split.

The second corpus variant, \ssfStratifiedCorpus (N=\ssfStratifiedCorpusSize), contains a flat sample of 45 stories from each of 50 subreddits, without any data splits, and is intended for analyses that require balanced representation across communities.

\subsection{Context Summarization}
To ground inferences in stories' social and conversational contexts, we summarize key community and conversational contexts and provide them as inputs to the inference generation task. 

\paragraph{Conversation Context}  We summarize the initial post along with up to 5 ancestral parent comments and 5 prior peer comments, approximating our assumed Reddit user browsing model of what contextual information a reader is likely to have seen (see App. \ref{app:reddit-browsing-model} for details). See App. \ref{app:prompt-templates-conversation-context} for the prompts.

\paragraph{Community Context} Conversations occur in distinct subreddits with varying purposes and norms, shaping what knowledge is assumed, what motivates participation, and how comments are interpreted. Drawing from a dataset of subreddits’ public self-descriptions and community guidelines \cite{Lloyd2025-rj}, we use GPT-4o to produce short summaries of the subreddits’ stated purpose and norms/values. See App. \ref{app:prompt-templates-community-context} for the prompts.

\subsection{Context Summary Validation}
Two authors evaluated the \textit{consistency} and \textit{relevance} of $N\geq30$ summaries for each source type (subreddit purpose, subreddit values/norms, initial post, conversation history) on a $5$-point Likert scale, following \citet{Fabbri2021-fa}. We observe moderately high inter-annotator agreement (IAA) and high summarization quality metrics, with mean scores $\ge4$. See App. \ref{app:ctx-sum-val} for definitions of the criteria, full annotation results, and qualitative error analyses highlighting outstanding challenges (e.g., reducing hallucination rate for very short contexts).

\section{\ssfFormalisms Modeling}
\label{sec:ssf-modeling}
We generate reference data and train open-weight models for inference generation and classification tasks, validated through human annotation.

\subsection{Inference Generation}
In preliminary qualitative investigations, GPT-4o outperformed the open-weight models within our compute budget for zero-shot inference generation. To balance performance with accessibility (reducing inference cost and compute requirements) and to support reproducibility, we adopt a model-distillation approach using SFT: we train a smaller open-weight student model on outputs generated by a stronger teacher model. We validate both the teacher and student models through large-scale human surveys.

\paragraph{Reference Data Generation}
To encourage diversity in the reference data, we prompt GPT-4o in a single API call to generate up to three independent inferences per (story, dimension) pair in \ssfSplitCorpus.
Each inference follows a dimension-specific template with one or more slots, designed to scaffold, without totally constraining, the free-text inferences. 
The template prefix primes the model to adopt the imagined perspective of community readers.
% many readers in the community.

\begin{figure*}[t]
    \centering    \includegraphics[width=\textwidth]{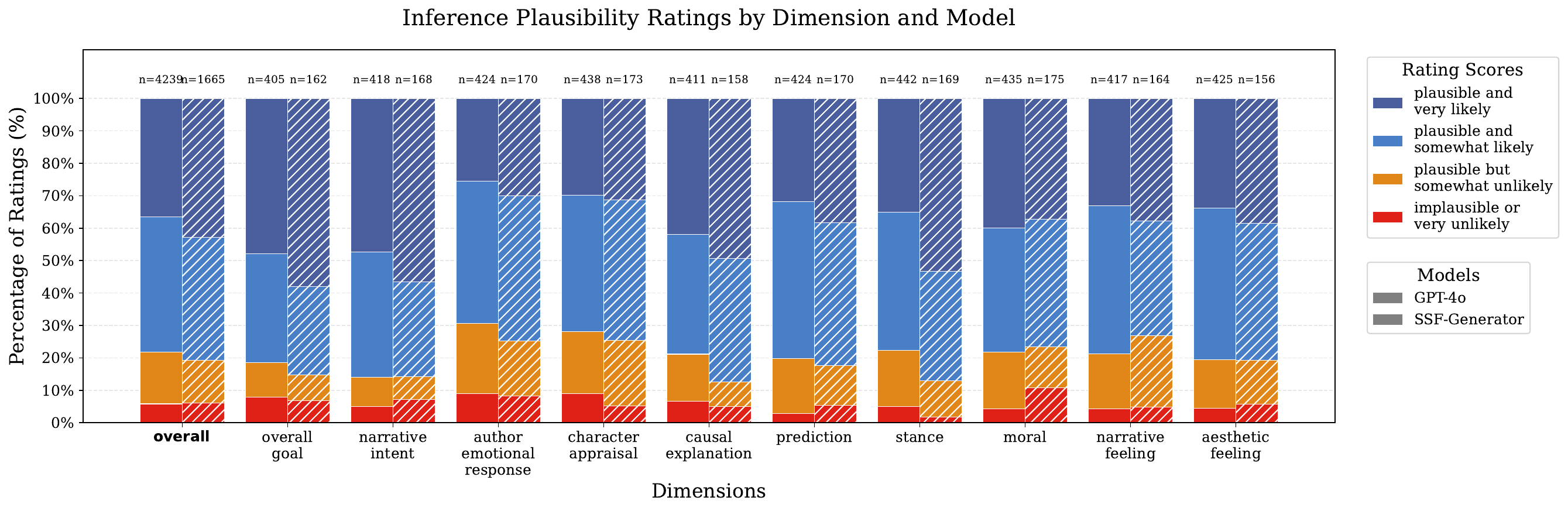}
    \caption{Human Inference Plausibility Ratings. The rates of ``very likely'' or ``somewhat likely''  indicate that GPT-4o and our distilled \ssfGenerator are effective at generating plausible inferences about reader response.} 
    \label{fig:plausibility_ratings_gpt4o}
\end{figure*}

\paragraph{\ssfGenerator Distillation}
We use LoRA \cite{Hu2021-kq} to finetune Llama3.1-8B-Instruct \cite{Dubey2024-jp} on the inferences generated by GPT-4o. 
% Due to its poorer instruction-following capabilities, we train \ssfClassifier to produce one inference at a time; however we still train on all GPT-4o outputs. 
Prompt templates are listed in App. \ref{app:prompt-templates-inference-generation}, and App. \ref{app-training-details} provides additional details on the distillation process, including generation parameters and finetuning configuration.

\paragraph{Validation}
Understanding what counts as a plausible reader response depends on perspective, particularly across communities. Since surveying members of $50+$ subreddits is logistically and financially onerous, we instead use a shared pool of human annotators, primed with rich community and conversational contexts to adopt the viewpoint of ``many readers from this community.''

We recruited a representative sample ($N{=}300$) of U.S. adults via Prolific to rate the contextual plausibility of GPT-4o-generated inferences, drawn from the test split of \ssfSplitCorpus. Using a POTATO-based annotation task \cite{Pei2022-yq}, participants read a storytelling comment in its community and conversational context and rated one random inference (of the $3$ possible) per taxonomy dimension for each of the $10$ dimensions on a $4$-point Likert scale based on prior work on commonsense plausibility evaluation \cite{Palta2024-gz, zhou-etal-2023-cobra}. 

To filter for annotation quality, we included a small percentage of \textit{known-implausible} inferences (see App. \ref{app:prolific-quality-filter} for details). $4{,}239$ ratings from $N{=}278$ annotators passed. A smaller validation survey for \ssfClassifier yielded $1{,}665$ ratings from $N{=}104$. See App. \ref{app:prolific-annotation-task} for additional details on the crowd annotation task, including recruitment materials and the annotation task interface. 

We report the survey results in Fig.~\ref{fig:plausibility_ratings_gpt4o}. Overall, $\geq 94\%$ of ratings were deemed plausible, and $\geq 78\%$ were deemed very or somewhat likely. This indicates that \ssfGenerator is fairly proficient at inferring probable reader response to social media stories across diverse contexts. See App. \ref{app:ssf-gen-err} for error analysis and App. \ref{app:ctx-ablations} for ablation experiments that remove community and/or conversational context during SFT distillation, demonstrating that incorporating context---especially conversational context---yields statistically significant improvements in \ssfGenerator’s alignment with its human-validated, full-context GPT-4o teacher.

\subsection{Inference Classification}
\paragraph{Reference Data Generation}
For multi-label inference classification, we find zero-shot performance to perform poorly overall, even with proprietary models, motivating the development of concrete annotation guidelines (see App. \ref{app:tax-ann-guide}) developed on a validation set as well as a $k$-shot prompting strategy that samples similar yet diverse examples via maximum marginal relevance \cite{Carbonell2017-oo}, among other methods. See App. \ref{app:k-shot-prompting-details} for $k$-shot sampling details and App. \ref{app:prompt-templates-inference-classification} for the prompt template.

\paragraph{\ssfClassifier Distillation}
In a similar fashion to the inference generation task, we use model distillation to train \ssfClassifier, which is Llama3.1-8B-Instruct finetuned to predict labels in a zero-shot fashion based on GPT-4.1 outputs. See App. \ref{app-training-details} for details.
\vspace{-0.25em}

\paragraph{Validation}
Two authors independently annotated a subset of $N=50$ test set examples for each dimension to assess inter-annotator agreement (IAA). We measured agreement on the multi-label classification task using the Jaccard Index and found strong overall consistency across dimensions (mean = 0.732; minimum $\geq 0.517$). After establishing reliability, the first author annotated $N=100$ validation and $N\geq100$ test examples following a protocol designed to (1) dynamically adjust annotation counts to mitigate label skew, and (2) identify extremely rare or nonexistent subdimension labels for exclusion from our models.

\begin{table*}[htbp]
\centering
\footnotesize
\resizebox{\textwidth}{!}{%
\begin{tabular}{l|r@{\hspace{2pt}}r|r@{\hspace{2pt}}r|r|r@{\hspace{2pt}}r|r@{\hspace{2pt}}r|r}
\hline
\multirow{3}{*}{Dimension} & \multicolumn{5}{c|}{Validation} & \multicolumn{5}{c}{Test} \\
\cline{2-11}
 & \multicolumn{2}{r|}{\refClassifier} & \multicolumn{2}{r|}{\ssfClassifier} & \multirow{2}{*}{N} & \multicolumn{2}{r|}{\refClassifier} & \multicolumn{2}{r|}{\ssfClassifier} & \multirow{2}{*}{N} \\
\cline{2-5}\cline{7-10}
 % & $F_1_{Micro}$ & $F_1_{Macro}$ & $F_1_{Micro}$ & $F_1_{Macro}$ & & $F_1_{Micro}$ & $F_1_{Macro}$ & $F_1_{Micro}$ & $F_1_{Macro}$ & \\
 & $F_{1_{\text{Micro}}}$ & $F_{1_{\text{Macro}}}$ & $F_{1_{\text{Micro}}}$ & $F_{1_{\text{Macro}}}$ & & $F_{1_{\text{Micro}}}$ & $F_{1_{\text{Macro}}}$ & $F_{1_{\text{Micro}}}$ & $F_{1_{\text{Macro}}}$ & \\

\hline
overall goal & $\gradient{0.84}$ & $\gradient{0.77}$ & $\gradient{0.87}$ & $\gradient{0.82}$ & $100$ & $\gradient{0.81}$ & $\gradient{0.77}$ & $\gradient{0.79}$ & $\gradient{0.73}$ & $100$ \\
narrative intent & $\gradient{0.80}$ & $\gradient{0.81}$ & $\gradient{0.82}$ & $\gradient{0.79}$ & $100$ & $\gradient{0.80}$ & $\gradient{0.69}$ & $\gradient{0.78}$ & $\gradient{0.73}$ & $101$ \\
author emotional response & $\gradient{0.95}$ & $\gradient{0.88}$ & $\gradient{0.94}$ & $\gradient{0.82}$ & $100$ & $\gradient{0.93}$ & $\gradient{0.86}$ & $\gradient{0.94}$ & $\gradient{0.84}$ & $297$ \\
character appraisal & $\gradient{1.00}$ & $\gradient{1.00}$ & $\gradient{1.00}$ & $\gradient{1.00}$ & $100$ & $\gradient{0.99}$ & $\gradient{0.99}$ & $\gradient{0.99}$ & $\gradient{0.99}$ & $259$ \\
causal explanation & $\gradient{0.81}$ & $\gradient{0.80}$ & $\gradient{0.78}$ & $\gradient{0.74}$ & $100$ & $\gradient{0.87}$ & $\gradient{0.85}$ & $\gradient{0.78}$ & $\gradient{0.74}$ & $135$ \\
prediction & $\gradient{0.90}$ & $\gradient{0.87}$ & $\gradient{0.87}$ & $\gradient{0.82}$ & $100$ & $\gradient{0.89}$ & $\gradient{0.82}$ & $\gradient{0.83}$ & $\gradient{0.73}$ & $117$ \\
stance & $\gradient{1.00}$ & $\gradient{1.00}$ & $\gradient{1.00}$ & $\gradient{1.00}$ & $100$ & $\gradient{1.00}$ & $\gradient{1.00}$ & $\gradient{1.00}$ & $\gradient{1.00}$ & $242$ \\
moral & $\gradient{0.80}$ & $\gradient{0.80}$ & $\gradient{0.72}$ & $\gradient{0.71}$ & $100$ & $\gradient{0.75}$ & $\gradient{0.72}$ & $\gradient{0.65}$ & $\gradient{0.62}$ & $100$ \\
narrative feeling & $\gradient{0.89}$ & $\gradient{0.78}$ & $\gradient{0.87}$ & $\gradient{0.67}$ & $100$ & $\gradient{0.88}$ & $\gradient{0.82}$ & $\gradient{0.85}$ & $\gradient{0.77}$ & $257$ \\
aesthetic feeling & $\gradient{0.94}$ & $\gradient{0.89}$ & $\gradient{0.95}$ & $\gradient{0.93}$ & $100$ & $\gradient{0.90}$ & $\gradient{0.86}$ & $\gradient{0.87}$ & $\gradient{0.76}$ & $156$ \\
\hline
\end{tabular}
} % end resizebox
\caption{Inference classification annotation results. \ssfClassifier approaches GPT-4.1-level performance.}
\label{tab:tax_class_results}
\end{table*}

Table \ref{tab:tax_class_results} reports the final annotation results, showing that the $k$-shot GPT-4.1 approach generalizes well to the test set, and that our finetuned zero-shot \ssfClassifier approaches GPT-4.1-level performance. \ssfClassifier exceeds, matches, or is within 0.05 F-1 points of GPT-4.1 for a majority of \ssfTaxonomy dimensions ($7/10$ for Micro F-1, $6/10$ Macro F-1), with all \ssfClassifier dimensions being within 0.1 points from their GPT-4.1 counterparts. See App. \ref{app:tax_class_validation} for the full dimension-level IAA results, additional annotation protocol details, and error analyses (e.g., challenges distinguishing informational support from persuasion).
\vspace{-0.5em}

\section{\ssfCorpus Analysis}
Next, we apply \ssfFormalisms to investigate narrative practices across Reddit communities.
\vspace{-0.25em}

\paragraph{Narrative Intents and their Relation to Overall Goals}
We examine how narrative intents are distributed across social media and how they connect to broader posting or commenting goals. The most common narrative intent is to justify or challenge a belief ($40\%$). Other common narrative intents include clarification ($14\%$), emotional release ($14\%$), showing one's identity ($10\%$), and entertaining ($10\%$). We display all dimension sublabel distributions in Fig. \ref{fig:all_freqs} in App. \ref{appendix-additional-results}.

Analyzing associations between dimensions, such as overall goal and narrative intents, reveals meaningful patterns. Using normalized pointwise mutual information (NPMI)  \cite{church_word_1990, bouma_normalized_2009} as a measure of association between taxonomy sublabels, we observe that the overall goal of providing emotional support is strongly associated with the narrative intent of conveying a similar experience (NPMI: $0.35$), which casts narrative as a crucial mechanism for empathy. See Fig. \ref{fig:npmi} in App. \ref{appendix-additional-results} for the full NPMI matrix). More broadly, combining distributional and association analyses enabled by \ssfFormalisms offers a systematic way to capture both common practices and subtler structural patterns.

\begin{figure}[t]
    \centering
    \includegraphics[width=\columnwidth]{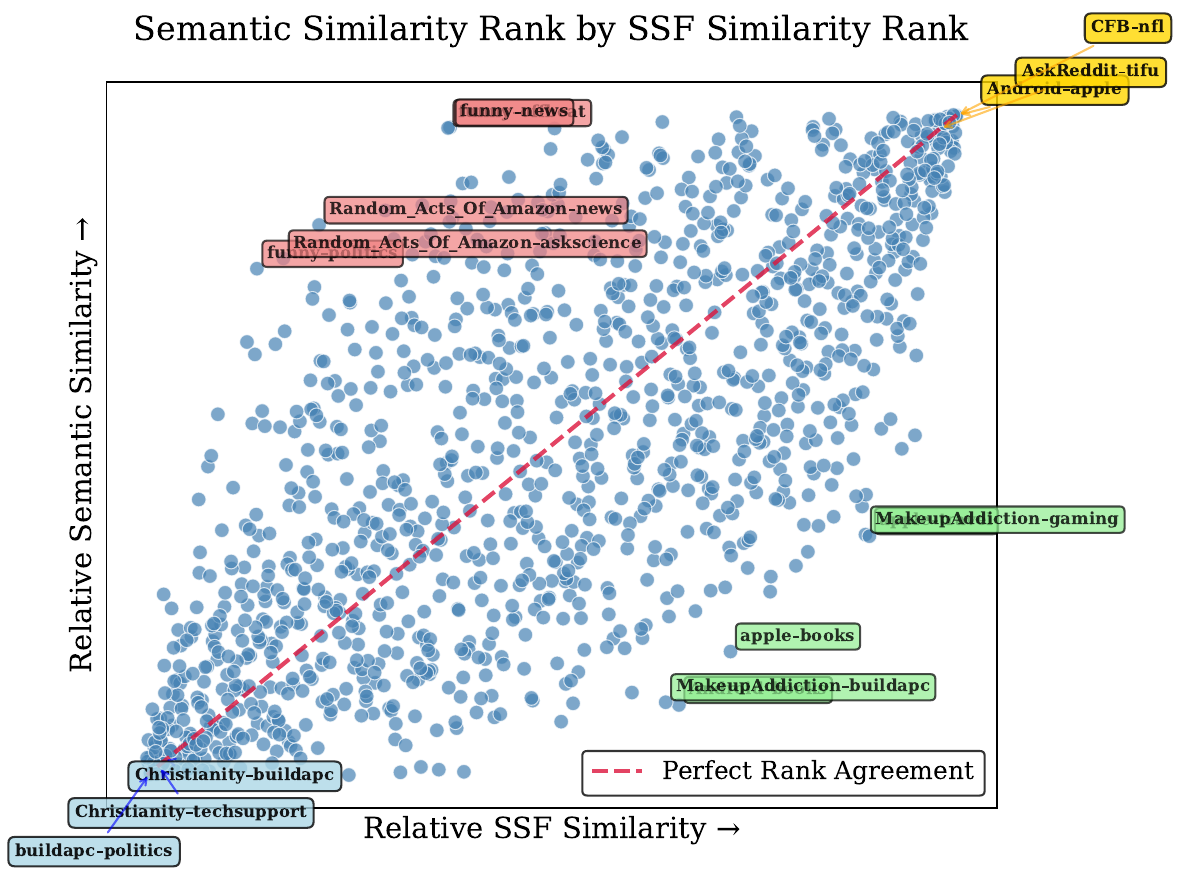}
    \caption{Subreddit similarity rankings according to semantic similarity vs \ssfSimMetric. Each point represents a subreddit pair, and off-diagonal points indicate metric disagreement.}
    \label{fig:similarity}
\end{figure}

\paragraph{Community Similarity} 
A key goal of our approach is to compare storytelling communities beyond surface-level topic overlap. Traditional semantic similarity misses how stories resemble each other in broader communicative, interpretive, and affective frames. To address this, we introduce \ssfSimMetric, a narrative similarity measure grounded in our \ssfFormalisms. Rather than comparing raw text embeddings, \ssfSimMetric computes similarity over (1) inferences generated by \ssfGenerator and (2) taxonomy label distributions predicted by \ssfClassifier, capturing extra-textual insights that transcend content.

While \ssfSimMetric draws on validated model outputs, we additionally assess its global construct validity via human annotation of story similarity with respect to communicative function and impact, capturing pragmatic and interpretive dimensions beyond surface semantics ($200$ stories; $N=50$ pairs of pairs). Inter-annotator agreement on a subset ($N=20$) was moderately high ($\kappa=0.5098$), and on the full set ($N=50$), \ssfSimMetric aligned with human judgments substantially more often than a Sentence-BERT \texttt{all-MiniLM-L6-v2} \cite{Reimers2019-ju} semantic similarity baseline (74\% vs. 52\%). See App. \ref{app:ssf-sim-details} for \ssfSimMetric’s formal definition and additional validation details.

Fig. \ref{fig:similarity} visualizes how subreddit-level similarity rankings diverge between semantic and narrative-reception-based measures. Where the two metrics agree, the results align with intuition: both identify dissimilarity between subreddit pairs such as \texttt{r/Christianity}-\texttt{r/buildapc} and \texttt{r/buildapc}-\texttt{r/politics}, while highlighting similarity between \texttt{r/Android}-\texttt{r/apple} and \texttt{r/AskReddit}-\texttt{r/tifu}. 

The divergences between metrics reveal the distinct contribution of our approach. Despite topical differences, our metric identifies strong similarity between pairs like \texttt{r/MakeupAddiction}-\texttt{r/buildapc} and \texttt{r/apple}-\texttt{r/books}, suggesting these communities engage in similar narrative practices regardless of subject matter. For example, one \texttt{r/MakeupAddiction} post seeking product reassurance---detailing brand, shade, and storage temperature---exhibits functional similarity to a \texttt{r/buildapc} post describing troubleshooting of voltage, GHz, and temperature specifications. 

In the upper-left quadrant of Fig. \ref{fig:similarity}, stories revolve around predictable topics yet function differently. For instance, the \texttt{r/funny} subreddit frequently engages with current events from \texttt{r/news} and \texttt{r/politics}, yet it adopts a markedly different communicative orientation---lighthearted rather than persuasive---yielding low \ssfSimMetric values despite topical overlap. For example, an \texttt{r/funny} post about a man gesticulating at a driver in a car registers as semantically similar but functionally dissimilar to a \texttt{r/news} story covering a problematic police car stop. See App. \ref{app:ssf-sim-vs-sem-examples} for these and other examples.

\paragraph{Community Narrative Diversity} 

\begin{figure}[t]
    \centering
    \includegraphics[width=\linewidth]{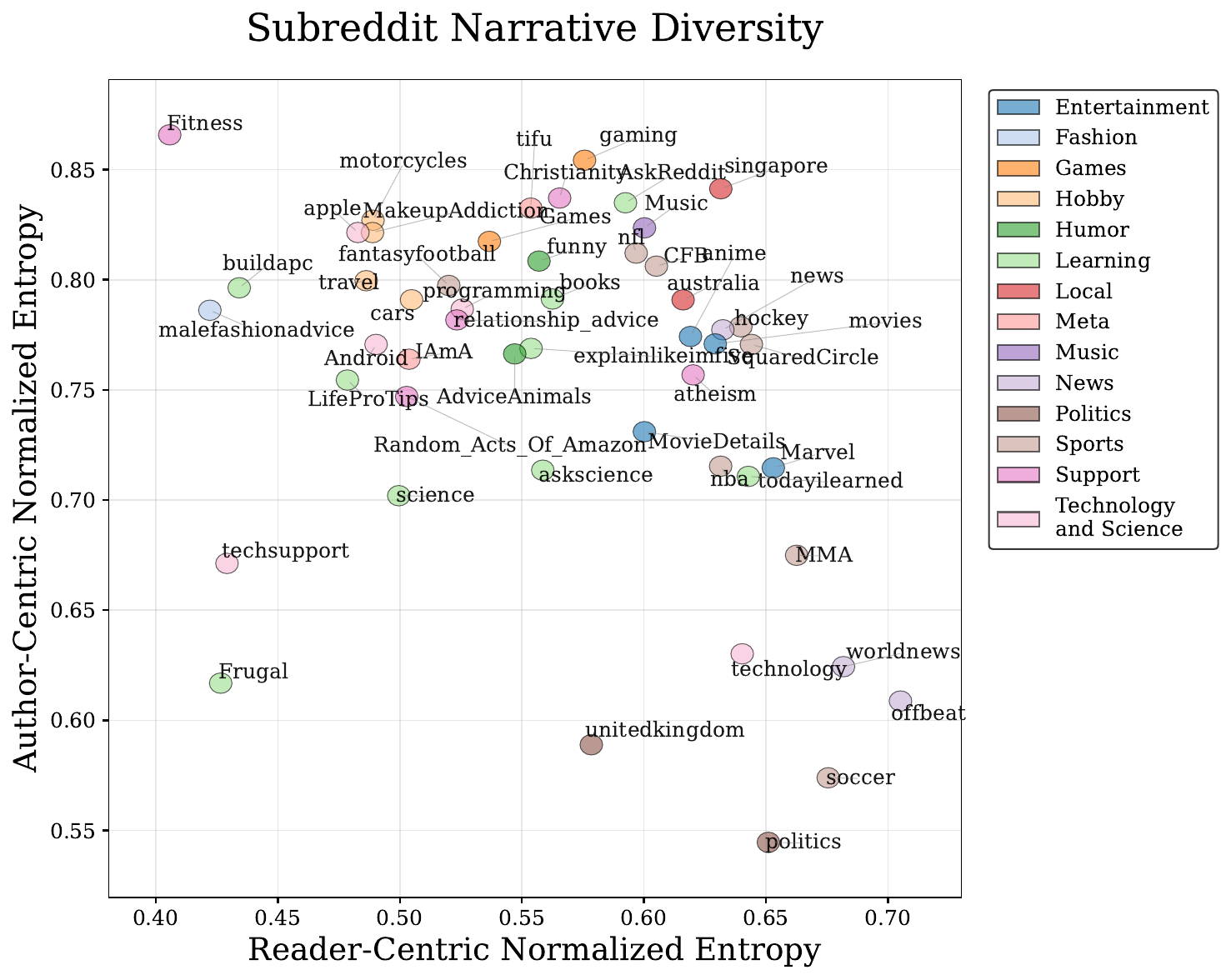}
    \caption{Author-centric and reader-centric lenses onto diversity of narrative practices across subreddits.}
    \label{fig:entropy_scatterplot}
\end{figure}

\ssfFormalisms enable analysis of community-level diversity in narrative intents and reader responses. Specifically, it can illuminate how author and reader behaviors vary in predictability and dynamism across communities, a phenomenon difficult to capture without a generic yet fine-grained framework. We measure diversity via normalized entropy of each \ssfTaxonomy dimension's subreddit-level sublabel distribution. Higher entropy indicates a wider variety of author goals, narrative strategies, or reader reactions. We partition the \ssfTaxonomy dimensions into two groups, \textit{author‑centric} (overall goal, narrative intent, author emotional response) and \textit{reader‑centric} (all others), then compute within-group normalized entropy mean across dimensions. We plot subreddits along these axes: the y-axis reflects variation in authorial intents and emotional responses, and the x-axis reflects variation in readers' interpretations and reactions (Fig. \ref{fig:entropy_scatterplot}).

The quadrants reveal distinct dynamics. \texttt{r/Frugal} and \texttt{r/techsupport} show low diversity on both axes, reflecting shared goals and procedural scripts. \texttt{r/Fitness} shows high author but low reader diversity, suggesting varied authorial approaches interpreted in predictable ways. Conversely, \texttt{r/worldnews}, \texttt{r/politics}, and \texttt{r/offbeat} demonstrate low author but high reader diversity, consistent with rapidly changing or contested content. We also observe topical trends:\footnote{We reuse subreddit topic labels from \cite{Fiesler2018-jr}.} support, gaming, and hobby communities skew toward author diversity, while sports, entertainment, and news communities show greater reader diversity (see Fig. \ref{fig:auth_entropy} in App \ref{appendix-additional-results}).

\section{Discussion}
\paragraph{Interpretive Communities}
Predicting reader response raises the problem of interpretive authority: whose understanding of a story counts?

While individual variation in narrative reception is well-documented \citep{Gerrig2022-ve}, our focus on contextual, community-specific reasoning aligns with theoretical perspectives in reader response and reception theory that emphasize commonalities across readers and shared interpretive frames in particular contexts \cite{Willis2017-lz, Mailloux1982-fg, Fish1990-fw}. \citeposs{Fish1990-fw} concept of interpretive communities---groups of readers with shared orientations and assumptions---serves as a point of departure for our exploration of community-specific patterns of commonsense reasoning about narrative reception. Our research design thus aimed to find a middle ground between assuming a universal response (disregarding context) vs. a highly individualized response.

Operationally, we modeled interpretive communities through subreddit affiliation, drawing on community guidelines and discussions in inference generation. Although we see clear trends using these community boundaries, our cross-community comparisons demonstrate that narrative practices are not strictly confined to subreddit boundaries. This points to a broader challenge: identifying the interpretive strategies and commitments that bind or distinguish individuals and groups without relying on predefined boundaries like subreddit labels. 

\paragraph{Toward Understanding the Social Functions of Storytelling}
Beyond NLP, a growing line of research has argued that reasoning about the social functions of narrative is crucial for connecting literary experience to broader societal questions \cite{Felski2014-xd, Dillon-SUnknown-qx}, and scholars across disciplines have proposed taxonomies from civic-minded \cite{Dillon-SUnknown-qx}, anthropological \cite{BoydUnknown-fm}, and psychological \cite{Walsh2022-sx} perspectives, among others. These efforts highlight storytelling's centrality in identity and sensemaking \citep{Holmes2005-lh, Somers1994-xm}. Research in psychology and cognitive science emphasizes narrative's unique affordances for social cognition: reading fiction supports theory-of-mind \citep{Mar2011-wj, Kidd2013-qx, Black2021-zq} and empathy \citep{Bal2013-ol, Koopman2015-yu, Keen2006-wa}, and facilitates vicarious experience through social simulation \citep{Oatley2016-xs, Tamir2016-px, Mar2008-pm}. Our work contributes to this growing body of research by demonstrating how NLP methods can be used to analyze large corpora of everyday storytelling and infer how stories function in specific social environments. Our bottom-up, inference-based approach complements both theoretical accounts and small-scale experimental studies, offering a scalable method for analyzing patterns of narrative use and interpretation across online social media contexts. 

\section{Conclusion and Future Work}
% \label{app:future-directions}
\ssfFormalisms illuminate the social topography of narrative practices and reader response in online communities, opening avenues for future research in computational social science and cultural analytics. Future work could explore:
\begin{itemize}[itemsep=0.5pt, topsep=2pt, left=1em]
    \item How narrative reception practices of communities and their members co-evolve.
    \item How individuals belonging to multiple interpretive communities navigate potentially contradictory commitments.
    \item Storytelling dynamics that foster prosocial outcomes in conversations, such as perspective-taking and learning.
    \item Personalization of models to individuals or communities, e.g., via Direct Preference Optimization \cite{Rafailov2024-kk} on pairwise preference data for inference plausibility.
\end{itemize}

By linking fine-grained, context-sensitive modeling with a generic taxonomy for reader response to social media storytelling, \ssfFormalisms enable nuanced, grounded studies of how narratives both shape and are shaped by online communities.

\section{Limitations}
\label{sec:limitations}
\paragraph{Community and Conversation Context} The iterative summarization strategy we used for summarizing conversational context, while efficient, can lead to cascading information loss at each step, resulting in non-optimal summarization quality. We also recognize the existence of alternative methods for richly representing community values and norms, e.g., \cite{Park2024-sj}, beyond our short summaries of subreddits’ self-written guidelines. 

\paragraph{Generalizability} The \ssfFormalisms formalism is not dependent on a particular platform, conversation structure, or topical domain, so we expect our framework to generalize fairly broadly across online conversations. However, there are important limitations. First, our instantiation of the framework in \ssfGenerator and \ssfClassifier on \ssfCorpus is restricted to English language conversations. Second, we do not expect our models to generalize to niche subreddits that require highly specialized domain knowledge. Third, we expect weaker generalization in communities where reader responses across members do not follow predictable patterns, whether due to high polarization or to largely idiosyncratic reactions among community members. In such settings, our framework---predicated on the existence of common or likely responses across many readers within a community---may not represent these communities well. Fourth, we filtered extremely toxic and sexually explicit content out of \ssfCorpus to prevent potential harm to annotators validating our models, accepting the tradeoff of decreased understanding of how our model handles this content type. Relatedly, our measurements for dimensions like stance and character appraisal may be slightly skewed toward supportive or positive judgments, due to less toxic content that might provoke negative reactions from readers. Finally, we do not model individuals, but rather groups of readers in particular communities and conversational contexts, as perceived by a representative sample of U.S. adults. 

\paragraph{Taxonomy} Our taxonomy is not exhaustive; it tries to balance breadth and depth. There is much work focused on particular dimensions that may yield more precise estimates of some narrow aspects of reader response, for example, the Story World Absorption Scale for measuring narrative absorption \cite{Kuijpers2014-gl}, although these survey methods do not necessarily transfer to the social media storytelling domain. Another example is the line of research in empirical literary studies on the complex interplay of different forms of feeling during literary reading that extends beyond discrete emotion taxonomies \cite{Miall2002-cj}.

\paragraph{Task Definition} While we take inspiration from discourse processing models like \citeposs{Graesser1994-gz} constructionist model as well as concepts from linguistic pragmatics like context selection \cite{Sperber1995-tr}, we do not explicitly model these cognitive processes. Instead, our models are designed to produce inferences that surpass a basic threshold of perceived plausibility and likelihood according to human annotators.

Moreover, in the absence of consensus on interdependency across reader response dimensions---an open question in cognitive psychology, empirical literary studies, and hermeneutics \cite{McNamara2009-ly, Bortolussi2002-gv, Pianzola2016-rz}---we assume that each dimension is independent and that readers generate at least one inference per dimension for each story. Theoretical and empirical inquiry into the dependencies between dimensions of reader response is an important direction for future work that may motivate joint, structured, or recursive modeling.

Finally, our plausibility rating task used for inference validation imposes a layer of indirection by priming respondents and annotators to consider the perspective of a contextually grounded group of readers. It is possible that this indirection could influence aspects of reader response in ways that are not well understood. 

\paragraph{Small-scale Global Validation of \ssfSimMetric} Although the \ssfSimMetric metric is composed of the extensively-validated outputs from \ssfGenerator and \ssfClassifier, we relied on comparatively fewer annotations focused on \textit{global} construct validity (App. \ref{app:ssf-sim-global-validation}).

\section{Ethical Considerations}
Our institution's IRB approved our study. Survey participants were paid at a $\$15/$hour rate.
When creating the \ssfCorpus, we masked PII and removed toxic and explicit texts before presenting the data to annotators. Following prior work \citep{bruckman2002studying}, we do not include any verbatim examples of personal stories in this paper.

It is important that research using the data of online communities not impede the goals of those communities \citep{zent2025beyond}, and we believe our work holds relatively low risk for our studied communities. As the goal of this paper was to draw comparisons across a large number of online communities, we could not engage deeply with the goals of each community, and while it was not feasible for us to engage users from each specific community, we conducted human evaluations that included important context about each community and its values and conversational dynamics. 

Misuse of our framework---intentional or otherwise---could produce representational harms, particularly if biases in either our models or our U.S.-based annotator pool are overlooked and implicitly treated as universal. Such misuse may lead researchers or practitioners to miss important cultural, gendered, or ideological variation across online communities. To mitigate these risks, we list several qualifications regarding the generalizability of our framework in Section \ref{sec:limitations}, and we require prospective users to acknowledge these limitations before accessing the models on HuggingFace.

\section*{Acknowledgments}
We thank our anonymous reviewers for their constructive feedback. We are also grateful to Patrick Park, Mingqian Zheng, and Chani Jung for comments on early versions of this work. This work was supported in part by the Block Center for Technology and Society at Carnegie Mellon University.

\bibliography{bib/custom}

\appendix

\pagebreak
\onecolumn

\section{\ssfTaxonomy}
\label{app:taxonomy} 
\begingroup
\scriptsize
\setlength{\LTcapwidth}{0.9\textwidth}
\begin{longtable}{p{0.2\textwidth}p{0.3\textwidth}p{0.4\textwidth}}
\caption{Taxonomy of pragmatic aspects of narrative intent and reception of storytelling on social media.}
\label{tab:taxonomy} \\
\toprule
\rowcolor[gray]{0.75}
\multicolumn{1}{p{0.2\textwidth}}{\textbf{Sub-dimension}} & \multicolumn{1}{p{0.3\textwidth}}{\textbf{Definition}} & \multicolumn{1}{p{0.4\textwidth}}{\textbf{Example}} \\
\midrule
\endfirsthead
\multicolumn{3}{p{0.9\textwidth}}{\tablename\ \thetable{} -- continued from previous page} \\
\toprule
\rowcolor[gray]{0.75}
\multicolumn{1}{p{0.2\textwidth}}{\textbf{Sub-dimension}} & \multicolumn{1}{p{0.3\textwidth}}{\textbf{Definition}} & \multicolumn{1}{p{0.4\textwidth}}{\textbf{Example}} \\
\midrule
\endhead
\midrule
\multicolumn{3}{r}{Continued on next page} \\
\endfoot
\bottomrule
\endlastfoot
\rowcolor{SkyBlue!15}
\multicolumn{3}{p{0.953\textwidth}}{
\textbf{Dimension: Overall Goal}\newline
\textbf{Template}: \textit{Many readers from this subreddit would think that the author's overall goal in posting/commenting this text was to \{\{SHORT VERB PHRASE DESCRIBING OVERALL GOAL\}\}.}\newline
\textbf{Summary}: Some common overall goals for posting on social media include:\par - request or provide facts or info about approaches, strategies, etc\par - seek or provide emotional support\par - express/reinforce one’s identity, values, or accomplishments\par - solicit or share experiences to gain or share insights\par - argue/advocate for a viewpoint to persuade others\par - share an enjoyable or funny post
} \\
\midrule
request\_info\_support & request factual info or advice about approaches, strategies, etc. & The author recounts a series of troubleshooting steps they've already followed to solve a technical issue, then poses a technical question. \\
provide\_info\_support & provide facts or advice about approaches, strategies, etc. & The author reports a newsworthy event relevant to the community. \\
request\_emotional\_support & express emotions or characterize a situation to elicit others' comfort, understanding, or empathy. & The author describes a series of unlucky events with their family and reports feeling misunderstood. \\
provide\_emotional\_support & provide emotional support to someone by acknowledging their identity, values, or accomplishments; or offering emotional comfort, understanding, or empathy. & The author justifies how someone who (presumably) posted a text earlier in the conversation reacted in some situation. \\
affirm\_identity\_self & reinforce or assert their own identity, values, or accomplishments. & The author asserts that they are a good student. \\
request\_experiential\_accounts*\footnote{Subdimensions marked with an asterisk were extremely rare in \ssfSplitCorpus and were thus excluded from \ssfClassifier modeling and validation. See App. \ref{app:tax-class-protocol} for additional context.} & solicit stories or experience from others to gain perspective or insight. & The author describes a strange experience and then asks if anyone else has experienced something similar. \\
provide\_experiential\_accounts & share a personal story or experience to inform or engage others. & The author shares a story about their immigration experience to show that it was different than how someone who (presumably) posted earlier characterized immigration. \\
persuade\_debate & advocate for a viewpoint or make an argument to convince others. & The author asserts that a sports team made a bad trade based on the poor track record of a player. \\
entertain & share an enjoyable or funny post. & The author repeats a famous joke from a comedian. \\
\midrule
\rowcolor{SkyBlue!15}
\multicolumn{3}{p{0.953\textwidth}}{
\textbf{Dimension: Narrative Intent}\newline
\textbf{Template}: \textit{Many readers from this subreddit would think that the author told the story in their post/comment to \{\{SHORT VERB PHRASE DESCRIBING NARRATIVE INTENT\}\}.}\newline
\textbf{Summary}: Some common reasons people tell a story in a social media post include:\par - express/demonstrate one’s identity or beliefs\par - teach a life lesson\par - justify or challenge a belief or social norm\par - entertain\par - release pent up emotions (vent)\par - convey need for emotional support\par - convey a similar experience\par - correct misunderstandings or add missing details associated with an event/situation
} \\
\midrule
show\_identity & demonstrate an aspect of their identity in action, by example. & The author shows that they can't be bought by telling a story about how they turned down a lucrative job offer that did not align with their values. \\
justify\_challenge\_offer\_belief\_norm & explain how they came to hold or question a belief or social norm, or to reinforce/demonstrate why it is correct, beneficial, misguided, or harmful; teach a life lesson or influence the behaviors or attitudes of readers & The author told a story about how they came to appreciate the politeness of small talk in the South in the U.S. \\
entertain & share an enjoyable or funny story. & The author tells a story about a dream where they were giving a presentation in their underwear. \\
release\_pent\_up\_emotions & express themself as a means of emotional release or sensemaking. & The author recounts being totally caught off guard in a Zoom call where they were fired unceremoniously. \\
convey\_emotional\_support\_need & describe events that have left them in an unfortunate or unresolved state, in need of emotional support. & The author tells a story about how they have become lonely over the last few months because there friends have stopped reaching out. \\
convey\_similar\_experience & share a story similar to another story or related to a situation under discussion. & The author tells a story about their experience with graduate school admissions in the context of a discussion about different people's similar experiences. \\
clarify\_what\_transpired & correct misunderstandings or add missing details associated with an event or situation under discussion. & The author points out that a discussion has overlooked several important factors or events that are necessary to understand a situation under discussion. \\
\midrule
\rowcolor{SkyBlue!15}
\multicolumn{3}{p{0.953\textwidth}}{
\textbf{Dimension: Author Emotional Response}\newline
\textbf{Template}: \textit{Many readers from this subreddit would think that telling the story in their post/comment would cause the author of the post/comment to feel \{\{SHORT NOUN PHRASE DESCRIBING EMOTION\}\}.}\newline
\textbf{Summary}: Some feelings an author could experience after telling a story in a social media post include: fear, guilt, anger, sadness, disgust, envy, joy, pride, relief, hope, compassion, appreciation, and connection.
} \\
\midrule
fear & a response to perceived danger or threat. & The author tells a story about how they heard banging on their door in the middle of the night. \\
guilt & remorse for violating personal or social standards. & The author tells a story about how they have been keeping the fact that they dropped out of college a secret from their parents. \\
anger & a strong reaction to perceived harm, injustice, or frustration. & The author tells a story about how their dog was attacked at a park by another dog whose owner did not use a leash. \\
sadness & a sense of loss, disappointment, or helplessness. & The author tells a story about how they were rejected from their dream school. \\
disgust* & a reaction of revulsion to something perceived as offensive or repellent. & The author tells a story about how they saw someone urinating on the subway. \\
envy* & a desire for something others have, coupled with resentment. & The author tells a story about how their coworker was promoted over them despite not being as good an employee as the author.  \\
joy & a state of happiness and contentment. & The author tells a story about their wedding night that they remember fondly. \\
pride & a sense of satisfaction from achievements or qualities. & The author tells a story about their coming out process and the ways their life has changed for the better since coming out. \\
relief & a release from stress or tension after resolving a concern. & The author tells a story about narrowly avoiding being caught skipping the last hour of their work shift. \\
hope & an optimistic expectation for a positive outcome. & The author tells a story about their persistence in trying to learn the piano and their excitement about improving despite setbacks. \\
compassion & concern for others' suffering. & The author tells a story about a limping dog they saw on the sidewalk on their way home from work. \\
appreciation & recognition and enjoyment of the good qualities of person, place, or thing. & The author tells a story about their gratitude for a mailman who delivered mail in their neighborhood for years. \\
connection & closeness or shared understanding with a person, place, or thing. & The author tells a story about how they have been building new friendships and building a new home in a city they recently moved to.  \\
\midrule
\rowcolor{SkyBlue!15}
\multicolumn{3}{p{0.953\textwidth}}{
\textbf{Dimension: Character Appraisal}\newline
\textbf{Template}: \textit{While or after reading the story within this text, many readers from this subreddit would \{\{EITHER "positively", "negatively", or "neutrally"\}\} judge \{\{EITHER "narrator" OR IDENTIFIER/NAME OF OTHER CHARACTER FROM STORY\}\}.}\newline
\textbf{Summary}: Readers often judge whether character’s actions or state (e.g., beliefs, values, goals). Some common types of character appraisals include: positive, negative, or neutral judgment of the narrator (or another character).
} \\
\midrule
positive\_appraisal\_narr & a positive judgment of the narrator's actions or state. & In the story within the post, the author intervened to help someone who was being harassed at a bar. \\
negative\_appraisal\_narr & a negative judgment of the narrator's actions or state. & In the story within the post, the author made a snide comment to a colleague. \\
neutral\_appraisal\_narr & a neutral appraisal of the narrator's actions or state. & In the story within the post, the author describes a routine day in their life as a student. \\
positive\_appraisal\_other\_char & a positive judgment of a non-narrator character's actions or state. & In the story within the post, the author's partner gave them a gift. \\
negative\_appraisal\_other\_char & a negative judgment of a non-narrator character's actions or state. & In the story within the post, someone proctoring a test that the author was taking ignored that some people were cheating. \\
neutral\_appraisal\_other\_narr & a neutral appraisal of a non-narrator character's actions or state. & In the story within the post, someone ordered a type of coffee that the narrator was unfamiliar with. \\
\midrule
\rowcolor{SkyBlue!15}
\multicolumn{3}{p{0.953\textwidth}}{
\textbf{Dimension: Causal Explanation}\newline
\textbf{Template}: \textit{While or after reading the story within the post/comment, many readers from this subreddit would think that \{\{SHORT DESCRIPTION OF SITUATION/STATE/ACTION FROM STORY\}\} could be explained by \{\{SHORT EXPLANATION\}\}.}\newline
\textbf{Summary}: Readers often make inferences to explain aspects of a story. Some common types of explanatory inferences readers make include:\par - inferences about the narrator’s (or other characters’) values, beliefs, or goals (to explain character actions or state)\par - inferences about why an event or state not directly caused by a human happened (e.g., why a natural disaster happened)
} \\
\midrule
narr\_explained\_by\_narr & explaining some aspect of the narrator (e.g., their feelings or behavior) based on the perceived state (e.g., underlying values, beliefs, or goals) or actions of the narrator. & In the story within the post, the narrator quits their job and moves to a new city. \\
narr\_explained\_by\_\\other\_char\_or\_thing & explaining some aspect of the narrator (e.g., their feelings or behavior) based on (1) the perceived state (e.g., underlying values, beliefs, or goals) or actions of a non-narrator character or (2) the perceived state of affairs or event not directly attributed to any character. & In the story within the post, the narrator expresses their frustration with a conversation they had with a friend. \\
other\_char\_or\_thing\_explained\_\\by\_narr & explaining some aspect of (1) a non-narrator character (e.g., their feelings or behavior) or (2) the perceived state of affairs or event not directly attributed to any character based on the perceived state (e.g., underlying values, beliefs, or goals) or actions of the narrator. & In the story within the post, the narrator describes their depression and how it may be impacting their partner. \\
other\_char\_or\_thing\_explained\_\\by\_other\_char\_or\_thing & explaining some aspect of (1) a non-narrator character (e.g., their feelings or behavior) or (2) the perceived state of affairs or event not directly attributed to any character on (a) the perceived state (e.g., underlying values, beliefs, or goals) or actions of a non-narrator character or (b) some other perceived state of affairs or event not directly attributed to any character. & In the story within the post, the narrator describes how their manager has been more lax during the COVID-19 pandemic. \\
\midrule
\rowcolor{SkyBlue!15}
\multicolumn{3}{p{0.953\textwidth}}{
\textbf{Dimension: Prediction}\newline
\textbf{Template}: \textit{While or after reading the story within the post/comment, many readers from this subreddit would predict that \{\{EITHER "the narrator" or NAME/IDENTIFIER OF OTHER CHARACTER OR THING FROM STORY\}\} might \{\{SHORT DESCRIPTION OF ACTION OR STATE\}\}.}\newline
\textbf{Summary}: Readers often predict what will happen next in the world of a story. Some common types of predictions include:\par - predictions about the future actions or state of the narrator (or other characters) values, beliefs, or goals (to explain character actions or mental state)\par - predictions about a future event or state not directly caused by a human (e.g., an asteroid hitting)
} \\
\midrule
narr\_future\_state & a prediction about the future state of the narrator. & The story within the post ends with the narrator's partner breaking up with the narrator. \\
narr\_future\_action & a prediction about the future actions of the narrator. & The story within the post ends with the narrator's partner making an ultimatum for the narrator. \\
other\_char\_future\_state & a prediction about the future state of a non-narrator character. & The story within the post ends with the narrator breaking up with their partner. \\
other\_char\_future\_action & a prediction about the future actions of a non-narrator character. & The story within the post ends with the narrator making an ultimatum for their partner. \\
non\_char\_thing\_future\_state* & a prediction about a future state not directly caused by a character  & The story within the post speculates about the state of technology in 100 years. \\
non\_char\_thing\_future\_event & a prediction about a future event not directly caused by a character  & The story within the post speculates on the impending demise of Black Friday sales due to increased online retail. \\
\midrule
\rowcolor{SkyBlue!15}
\multicolumn{3}{p{0.953\textwidth}}{
\textbf{Dimension: Stance}\newline
\textbf{Template}: \textit{After reading the story within this text, many readers from this subreddit would \{\{EITHER "support", "counter", or "be neutral to"\}\} the author's opinion\{\{SHORT DESCRIPTION OF AUTHOR'S OPINION/STANCE\}\}.}\newline
\textbf{Summary}: Readers often take a stance or overall opinion in response to a main idea, argument, or point advocated for by the author of a story, implicitly or explicitly. Readers can support (agree with), counter (disagree with), or be neutral to the author’s stance.
} \\
\midrule
support\_belief\_norm & a stance that mostly agrees with the stance most recently expressed (explicitly or implicitly) in the preceding conversation.  & The story within the post is about how the narrator conducted extensive research to learn that the risk of shark attacks on the beaches of California is marginal. \\
counter\_belief\_norm & a stance that mostly disagrees with the stance most recently expressed (explicitly or implicitly) in the preceding conversation.  & The story within the post is about how gambling is an effective way to get out of debt. \\
neutral\_belief\_norm & a stance that is neutral to the stance most recently expressed (explicitly or implicitly) in the preceding conversation.  & The story within the post is about the narrator's strong opinion about a highly niche topic that most people don't know or care about. \\
\midrule
\rowcolor{SkyBlue!15}
\multicolumn{3}{p{0.953\textwidth}}{
\textbf{Dimension: Moral\footnote{Although we perform modeling based on these moral subdimensions, we aggregate them into their higher-level categories following \cite{Schwartz2012-xk} for validation and analyses. The mapping is defined as follows: self-enhancement (achievement, power), openness to change (stimulation, self-direction), conservation (security, conformity, tradition), self-transcendence (universalism, benevolence), and hedonism.}}\newline
\textbf{Template}: \textit{While or after reading the story within the post/comment, many readers from this subreddit would think that the moral of the story is \{\{MORAL/THEME\}\}.}\newline
\textbf{Summary}: Some common morals/values that stories highlight the importance of include: independence, novelty/change, pleasure, achievement, power, security, conformity, tradition, benevolence, and universalism.
} \\
\midrule
self-direction & independent thought and action, expressed in choosing, creating and exploring. & The story within the text is about how Copernicus discovered that the Earth revolves around the Sun despite the contemporary belief that the Earth was at the center of the solar system. \\
stimulation & excitement, novelty, and challenge in life. & The story within the text is about someone travelling internationally for the first time. \\
hedonism & pleasure, enjoyment, or sensuous gratification for oneself. & The story within the text is about how someone had an exhilarating experience seeing their favorite band at a concert.  \\
achievement & personal success through demonstrating competence according to social standards & The story within the text is about a person earning their nursing degree. \\
power & control or dominance over people and resources; social status and prestige & The story within the text is about how a new CEO changed the direction of a company and let go 200 employees. \\
security & safety, harmony, and stability of society, of relationships, and of self. & The story within the text is about how a narrator had a difficult conversation with their mother about their relationship dynamic which benefited it in the long run. \\
conformity & restraint of actions, inclinations, and impulses likely to upset or harm others and violate social expectations or norms in everyday interactions, usually with close others & The story within the text is about how the narrator navigated flirtatious conversations with a coworker when the narrator was in a committed relationship. \\
tradition & respect, commitment, and acceptance of the customs and ideas that one's culture or religion provides. & The story within the text is about how someone's religious beliefs and practices evolved over time. \\
benevolence & preserving and enhancing the welfare of those with whom one is in frequent personal contact (the 'in-group'). & The story within the text is about how the narrator's neighbors brought him food each night while he recovered from a surgery. \\
universalism & understanding, appreciation, tolerance, and protection for the welfare of all people and for nature. & The story within the text is about how people everywhere are struggling due to labor issues with the rise of automation. \\
\midrule
\rowcolor{SkyBlue!15}
\multicolumn{3}{p{0.953\textwidth}}{
\textbf{Dimension: Narrative Feeling}\newline
\textbf{Template}: \textit{While or after reading the story within the post/comment, the narrative content (i.e., the characters' situation and events) would spur many readers from this subreddit to feel \{\{FEELING/EMOTION\}\}.}\newline
\textbf{Summary}: Readers may sympathize or empathize with emotions of the narrator or other characters in a story, or they may feel emotions that differ from emotions of characters. The content of stories can also evoke memories in readers, causing readers to remember or relive emotions. Some of the feelings a reader could experience while or after reading a story include: fear, guilt, anger, sadness, disgust, envy, joy, pride, relief, hope, compassion, appreciation, and connection.
} \\
\midrule
fear & a response to perceived danger or threat. & The story within the post is about increased violence in their community. \\
guilt* & remorse for violating personal or social standards. & The story within the post is about how the community (of which the reader is presumably a part) has failed to uphold their community standards. \\
anger & a strong reaction to perceived harm, injustice, or frustration. & The story within the post is an attack on a marginalized group. \\
sadness & a sense of loss, disappointment, or helplessness. & The story within the post is about the difficulty of making friends in a new city. \\
disgust & a reaction of revulsion to something perceived as offensive or repellent. & The story within the post is about a time the author accidentally sneezed into a customer's food while working at a restaurant and didn't tell them. \\
envy* & a desire for something others have, coupled with resentment. & The story within the post is about the author winning the lottery. \\
joy & a state of happiness and contentment. & The story within the post depicts a heartwarming moment about two twins separated at birth reunited after 20 years. \\
pride & a sense of satisfaction from achievements or qualities. & The story within the post is about a community to which both the author and reader belong rebuilt itself after a terrible flood. \\
relief* & a release from stress or tension after resolving a concern. & The story within the post is about how the author almost fell from a rock face during a free climb, but managed to save themselves. \\
hope & an optimistic expectation for a positive outcome. & The story within the post in a job forum is about how a company just posted 100 new job listings. \\
compassion & sympathy and concern for others' suffering. & The story within the post is about a fictional anthropomorphic mouse that got caught in a mouse trap. \\
appreciation & recognition and enjoyment of the good qualities of person, place, or thing. & The story within the post is about how the author has regularly picked up litter in a local creek for over a decade. \\
connection & closeness or shared understanding with a person, place, or thing. & The story within the post in a forum dedicated to a particular U.S. State recounts how the governor has stood up for the state in national politics. \\
\midrule
\rowcolor{SkyBlue!15}
\multicolumn{3}{p{0.953\textwidth}}{
\textbf{Dimension: Aesthetic Feeling}\newline
\textbf{Template}: \textit{While or after reading the story within the post/comment, the narrative form/techniques such as \{\{BRIEF DESCRIPTION OF TECHNIQUE OR FORMAL ELEMENT\}\} would spur many readers from this subreddit to feel \{\{FEELING\}\}.}\newline
\textbf{Summary}: The form, techniques, or style of a story can evoke aesthetic feelings in readers. Some aesthetic feelings a reader might experience include: suspense (anticipation for imminent event), curiosity (desire for information from the past to explain the present), surprise (experiencing a shocking/unexpected event), attentional engagement (feeling attention is grabbed and held by the story), feeling pulled into the world of the story, and visualization of images.
} \\
\midrule
suspense & a feeling of excitement or anxiety in anticipation of an imminent event. & The story within the text recounts the narrator's anxious week at work as they anticipated getting let go after hearing about company-wide layoffs.  \\
curiosity & desire for information from the past to explain the present & The story within the text introduces a character with a stab wound, but does not say how they were wounded. \\
surprise & experiencing an unexpected or shocking event. & The story within the text describes the narrator accidentally cutting the tip of their finger off while making dinner. \\
attention\_engagement & finding the story to be compelling and capable of holding one's attention, as opposed to mundane or banal. & The story within the text describes a high-speed getaway from a bank robbery. \\
transportation & immersion, absorption, or feeling pulled into the world of the story. & The story within the text is suspenseful and describes how the narrator was running from someone chasing them late at night in the woods. \\
evocation & visualization, e.g., due to vivid language. & The story contains a whimsical description of the variously shaped clouds floating through the sky. \\
amusement & finding the story to be funny or to contain amusing elements. & The story within the text recounts the narrator mistaking their best friend for their romantic partner. \\
other & an aesthetic feeling not covered by the provided categories & \textasciitilde{} \\
\midrule
\end{longtable}
\endgroup
\twocolumn

\section{Background: The Social Functions of Storytelling}
\label{app:background-social-functions}

Some theorists resist an instrumental view of storytelling, rejecting efforts to taxonomize literary experience and instead emphasizing fiction’s non-instrumental value. Others, however, argue that such systematization is useful for connecting storytelling with broader social issues and research \cite{Felski2014-xd, Dillon-SUnknown-qx}. 

Literary theorist \citet{Felski2014-xd} advocates for a critical openness to reasoning about the various (albeit overlapping) social functions of storytelling. The move toward taxonomizing, in her view, can offer a lens into ``aspects of reading that have suffered the repeated ignominy of cursory or cavalier treatment''. Particularly in the context of short, informal, everyday stories on online forums, as opposed to long-form literary fiction, understanding simple stories in terms of perceived intent and social impact offers an essential lens on communication and identity in online communities.

\citet{Felski2014-xd} proposes four modes of textual engagement, in particular: through reading stories, we learn about ourselves (recognition), have aesthetic experiences (enchantment), encounter configurations of social knowledge, and experience shock---all of which contribute to our capacity to develop and transform over time. Scholars from other disciplines have proposed alternative taxonomies tailored to particular aspects of literary impact at individual and social levels. \citet{Dillon-SUnknown-qx}, advocating for greater attention to narrative evidence as a crucial yet overlooked input to policymaking, propose four ``cognitive and collective'' functions of storytelling. As their argument goes, stories broaden perspective (by foregrounding alternative points of view), construct and cohere collective identities, represent models of the world, and generate anticipations for the future. From an evolutionary perspective, \cite{BoydUnknown-fm} describes storytelling as an adaptive function that promotes social cognition, coordination, and creativity. Psychologists \citet{Walsh2022-sx} propose three social functions: social learning (through indirect experience and increased memorability of stories), persuasion (e.g., due to decreased reactance), and coordination (through establishing common knowledge and consensus around actors, explanations, predictions, etc).

Thematically, these taxonomizing efforts, as well as more targeted research concerned with a particular social function (e.g., argumentation), point to the centrality of stories for identity and sensemaking. More than a mere tool for communicating identity, narrative is a fundamental mode through which we form identity \cite{Holmes2005-lh, Pellauer2022-mq}, an ``ontological condition of social life'' \cite{Somers1994-xm}. Furthermore, narrative cognition is how we construct our social reality, e.g., by reasoning about intentional agents and time \cite{Bruner1991-ny}. We socially negotiate and create meaning through criteria such as narrative coherence and fidelity of a story to our lived experiences \cite{Fisher1984-kw, Bietti2019-vp}.

Narrative sensemaking is especially suited to social understanding and learning. Stories engage readers in social simulation of narrative models of the world \cite{Oatley2016-xs, Tamir2016-px, Zunshine2006-yt, Mar2008-pm}, affording a kind of vicarious experience that can enhance social cognition \cite{Dodell-Feder2018-fr}. Specifically, reading fiction has been shown to enhance theory-of-mind \cite{Black2021-zq, Kidd2013-qx, Mar2011-wj} and empathy \cite{Bal2013-ol, Koopman2015-yu, Keen2006-wa}.

Additionally, stories play important roles in persuasion \cite{Green2000-nv, Falk2022-et}, public argumentation and community coalescence \cite{Pera2024-nv, Hou2023-fx, Polletta2006-lw}, and disclosure in institutional processes \cite{Polletta2011-kp}.

In our work, we approach the question of the social function of storytelling in a bottom-up fashion by focusing on the types of inferences and reactions readers make while or after reading stories. Engaging readers with questions of perceived intent and likely interpretations or emotional responses helps concretize the multi-dimensionality of narrative reception, offering a lens onto the broader interpretative, affective, evaluative, and ultimately social dynamics surrounding storytelling across different contexts. 

\section{Prompt Templates}
\label{app:prompt-templates}

\subsection{Conversation Context Summarization}
\label{app:prompt-templates-conversation-context}

\subsubsection{Post/comment summarization}
\begin{tcolorbox}[
  colback=gray!5!white,
  colframe=gray!75!black,
  title=Prompt Template (\textnormal{GPT-4o; T=0}),
  breakable
]
\small
The following text comes from a social media forum. Summarize the text in a maximum of 2 sentences. Do not hallucinate and do not say that the text is too short to summarize. \\

<<TEXT>>
\normalsize
\end{tcolorbox}

\subsubsection{Initial Post Summary}
\begin{tcolorbox}[
  colback=gray!5!white,
  colframe=gray!75!black,
  title=Prompt Template (\textnormal{GPT-4o; T=0}),
  breakable
]
\small
Your task is to distill the provided context about the top-level post in a subreddit conversation into a succinct 1-sentence summary.
\vspace{1.5em}

Context Types: \\
- Top-level Post Title: the title of the initial top-level post \\
- Top-level Post Summary: a summary of the initial top-level post \\

Context: \\
- Top-level Post Title: <<TITLE>> \\\
- Top-level Post Summary: <<POST\_SUMMARY>> \\

Write a 1-sentence summary of the provided context. Output just the summary and no other text. Start your response with 'The first post...'.
\end{tcolorbox}
\normalsize

\subsubsection{Ancestral Chain Summarization}
\begin{tcolorbox}[
  colback=gray!5!white,
  colframe=gray!75!black,
  title=Prompt Template (\textnormal{GPT-4o; T=0}),
  breakable
]
\small
Below are <<SUMMARY\_COUNT>> summaries social media posts in an ancestral chain (parent-child relationships as you read left to right). Your task is to generate a global summary of the overall chain based on the local summaries in three sentences or less. \\
- <<FIRST\_SUMMARY>> \\
... \\
- <<LAST\_SUMMARY>>
\normalsize
\end{tcolorbox}

\subsubsection{Preceding Peer Chain Summarization}
\begin{tcolorbox}[
  colback=gray!5!white,
  colframe=gray!75!black,
  title=Prompt Template (\textnormal{GPT-4o; T=0}),
  breakable
]
\small
Below are <<SUMMARY\_COUNT>> summaries of a chain of social media comments under a single parent post/comment. Your task is to generate a global summary of the overall chain based on the local summaries in three sentences or less. \\
- <<FIRST\_SUMMARY>> \\
... \\
- <<LAST\_SUMMARY>>
\normalsize
\end{tcolorbox}

\subsubsection{Conversation Context Summarization}
\begin{tcolorbox}[
  colback=gray!5!white,
  colframe=gray!75!black,
  title=Prompt Template (\textnormal{GPT-4o; T=0}),
  breakable
]
\small
Your task is to distill the provided context about the conversational context into a 1-2 sentence summary. \\

Conversational Context Types: \\
- Ancestors Summary: a summary of the chain of texts formed by a parent-child relationship leading up to the current text \\
- Preceding Peers Summary: a summary of the chronologically-ordered comments preceding the current text under the same parent \\

Conversational Context: \\
- Ancestors Summary: \\<<ANCESTRAL\_CHAIN\_SUMMARY>> \\
- Preceding Peers Summary: \\<<PRECEEDING\_PEERS\_SUMMARY>> \\

Summarize the provided conversational context in 1-3 sentences. Output just the summary and no other text. Start your response with 'The conversation so far...'.
\normalsize
\end{tcolorbox}

\subsection{Community Context Summarization}
\label{app:prompt-templates-community-context}

\subsubsection{Subreddit Purpose Summarization}
\begin{tcolorbox}[
  colback=gray!5!white,
  colframe=gray!75!black,
  title=Prompt Template (\textnormal{GPT-4o; T=0}),
  breakable
]
\small
Summarize the following description of the r/<<SUBREDDIT\_NAME>> subreddit in 1 sentence. Do not hallucinate and do not say the text is too short to summarize. Output the summary and no other text. \\

<<SUBREDDIT\_PUBLIC\_DESCRIPTION>>
\normalsize
\end{tcolorbox}

\subsubsection{Subreddit Values/Norms Summarization}
\begin{tcolorbox}[
  colback=gray!5!white,
  colframe=gray!75!black,
  title=Prompt Template (\textnormal{GPT-4o; T=0}),
  breakable
]
\small
Summarize key values or norms of the r/<<SUBREDDIT\_NAME>> subreddit that are either explicitly stated or strongly evidenced by the following description and rules for the subreddit. Do not hallucinate. Output a 1 sentence summary and no other text. \\

Description: \\<<SUBREDDIT\_DESCRIPTION\_SUMMARY>> \\
Rules: \\<<SUBREDDIT\_PUBLIC\_RULES>>
\normalsize
\end{tcolorbox}

\subsection{Inference Generation}
\label{app:prompt-templates-inference-generation}

\begin{tcolorbox}[
  colback=gray!5!white,
  colframe=gray!75!black,
  title=Prompt Template (\textnormal{GPT-4o; T=0}),
  breakable
]
\small
Your task is to use commonsense to generate one contextually plausible description of the <<TAXONOMY\_DIMENSION>> in a social media conversation. \\

General (non-exhaustive) information to help scaffold your thinking about <<DIMENSION>> in the context of social media storytelling:
<<DIMENSION\_OVERVIEW>> \\

The following conversational context types are available: \\
- Subreddit Name: the Reddit community where the conversation takes place \\
- Subreddit Description: a brief overview of the subreddit topic \\
- Subreddit Values: a high-level summary of key values, norms, or rules in the subreddit \\
- Top-level Post Summary: a summary of the first, top-level post in the conversation thread \\
- Conversation Summary: a summary of the prior conversation leading up to the current text \\
- Current Text: the current text to analyze. The text necessarily contains storytelling (even if the story is short or banal). \\

Conversational Context: \\
- Subreddit Name: \\<<SUBREDDIT\_NAME>> \\
- Subreddit Description: \\<<SUBREDDIT\_DESCRIPTION>> \\
- Subreddit Values: \\<<SUBREDDIT\_VALUES>> \\
- Top-level Post Summary: \\<<INITIAL\_POST\_SUMMARY>> \\
- Conversation Summary: \\<<CONVERSATION\_SUMMARY>> \\

Current Text: \\
<<CURRENT\_TEXT>> \\

Output Instructions: \\
Remember: Your task is to use commonsense to generate one contextually plausible description of the <<TAXONOMY\_DIMENSION>> in a social media conversation.
You may use the provided info about <<DIM>> as background but do **not** force your response to fit it. You must not copy directly from the provided info if you can answer more precisely in your own words. \\

**IMPORTANT RULES (READ CAREFULLY):** \\
- ONLY edit inside double-braced placeholders like `\{\{...\}\}`. DO NOT MODIFY ANY TEXT OUTSIDE `\{\{\}\}`. \\
- DO NOT change or correct the template’s wording, punctuation, or singular/plural mismatches. FOLLOW THE TEMPLATE EXACTLY. \\
- DO NOT modify the JSON structure. Use valid JSON with double quotes only. \\
- OUTPUT ONLY the completed template below — NO EXTRA TEXT, HEADINGS, OR COMMENTS. \\
- IF YOU BREAK THESE RULES, THE OUTPUT WILL BE UNUSABLE. \\
\{"response": <<DIMENSION\_TEMPLATE>>\}
\normalsize
\end{tcolorbox}

\subsection{Inference Generation - Known Implausible}
\label{app:prompt-templates-inference-generation-implausible}
The prompt follows a similar structure as in the standard case (see App. \ref{app:prompt-templates-inference-generation}), except the task instruction is:
\begin{tcolorbox}[
  colback=gray!5!white,
  colframe=gray!75!black,
  title=Prompt Template (Excerpt),
  breakable
]
\small
Your task is to generate one contextually **implausible** description of the TAXONOMY\_DIMENSION in a social media conversation.
\normalsize
\end{tcolorbox}

Additionally, the output format instructions clarify that the generated inferences will be used to test annotators: 
\begin{tcolorbox}[colback=gray!5!white,colframe=gray!75!black,title=Prompt Template (Excerpt),breakable]
\small
Your output will be used to test human annotators, to see if they correctly identify your response as implausible or extremely unlikely. 
\normalsize
\end{tcolorbox}

\subsection{Inference Classification}
\label{app:prompt-templates-inference-classification}
\begin{tcolorbox}[
  colback=gray!5!white,
  colframe=gray!75!black,
  title=Prompt Template (\textnormal{GPT-4.1; T=0}),
  breakable
]
\small
Using the taxonomy and tips below, classify the following description of the <<DIMENSION>> in a social media conversation. \\

Taxonomy for <<DIMENSION>>: \\
- <SUBDIM\_1>: <SUBDIM\_1\_DEF> \\
- <SUBDIM\_2>: <SUBDIM\_2\_DEF> \\ 
... \\
- <SUBDIM\_N>: <SUBDIM\_N\_DEF> \\

Classification Tips for <<DIMENSION>>: \\
<<DIMENSION\_ANNOTATION\_GUIDELINES>> \\

Examples: \\
Input: INFERENCE\_1 \\
Output: \{"response": \\<<GOLD\_SUBDIM\_LABELS\_FOR\_INF\_1>>\} \\

Input: INFERENCE\_2 \\
Output: \{"response": \\<<GOLD\_SUBDIM\_LABELS\_FOR\_INF\_2>>\} \\
... \\
Input: INFERENCE\_K \\
Output: \{"response": \\<<GOLD\_SUBDIM\_LABELS\_FOR\_INF\_K>>\} \\

Text to classify: \\
<<TEXT\_TO\_CLASSIFY>> \\
        
Output Instructions: \\
Remember: Using the taxonomy and tips below, classify the following description of the <<DIMENSION>> in a social media conversation. \\
Fill in the JSON list below with *ALL* of the categories that apply to the text. Many texts span multiple categories—please include every one that applies, not just the most obvious.' \\

**IMPORTANT RULES (READ CAREFULLY):** \\
- DO NOT modify the JSON structure. Use valid JSON with double quotes only. \\
- OUTPUT ONLY the completed template below — NO EXTRA TEXT, HEADINGS, OR COMMENTS. \\
\{"response": ["category\_a", "..."]\}
\normalsize
\end{tcolorbox}

The ``Examples'' section is used for reference data generation. For \ssfClassifier training and subsequent inference, the ``Examples'' section is excluded.

\section{Dataset Details}
\subsection{\ssfCorpus Curation}
\label{appendix-dataset-curation}

To curate \ssfCorpus, we preprocess and filter \textsc{reddit-corpus-small} in four steps. First, we use ConvoKit's TextCleaner to mask lingering PII. Second, we apply StorySeeker \cite{Antoniak2024-ko}, a binary story classifier trained on Reddit data, filtering for texts with a predicted story probability $\geq 0.7$ and a minimum length of 175 characters—thresholds chosen to reduce the elevated false positive rate observed with lower cutoffs. Third, we use the Perspective API to flag and remove posts and comments labeled \textsc{TOXICITY} or \textsc{SEXUALLY-EXPLICIT} (probability $\geq 0.5$), to mitigate the risk of exposing annotators to harmful content.\footnote{\url{https://developers.perspectiveapi.com/s/about-the-api-attributes-and-languages?language=en\_US}} Finally, we exclude subreddits that are primarily image-based, narrowly centered on niche cultural artifacts (e.g., a single video game), or consistently toxic or explicit.

\subsection{Formal Dataset Structure}
\label{appendix-dataset-structure}
Let \(G=(V,E)\) be a directed acyclic graph (DAG) representing a conversation where:

\begin{itemize}[itemsep=0.5pt, topsep=2pt, left=1em]
    \item \(V\) is the set of utterances, each composed of a \texttt{text} and \texttt{timestamp}.
    \item \(E \subseteq V \times V \times T\) is the set of typed directed edges, where $(v_i,v_j,t) \in E$ denotes an edge from $v_i$ to $v_j$ of type \(t \in \{par, pre\}\) corresponding to parent and preceding peer edges.
    \item There are no cycles, i.e. there is no directed path $(v_1,\ldots,v_k)$ where $v_1=v_k$.
    \item Each utterance \(v_j \in V\) may have at most one parent edge  \((v_i, v_j, par) \in E\) and one preceding peer edge \((v_k, v_j, pre) \in E\).
\end{itemize}

We recursively define the following functions for retrieving an utterance's ancestors (\textsc{ancs}) and chronologically-preceding peers (\textsc{pres}), respectively:

\small
\begin{math}
\begin{aligned}
    \textsc{ancs}(v_j) &= \begin{cases} 
     [] & \nexists \, (v_i, v_j, par) \in E \\[0.5em]
     \textsc{ancs}(v_i) + [v_i] & \exists \, (v_i, v_j, par) \in E
     \end{cases} \\[1em]
    \textsc{pres}(v_j) &= 
     \begin{cases} 
     [] & \nexists \, (v_k, v_j, pre) \in E \\[0.5em]
     \textsc{pres}(v_k) + [v_k] & \exists \, (v_k, v_j, pre) \in E
     \end{cases}
\end{aligned}    
\end{math}
\normalsize

\subsection{Defining Conversational Context: Assumed Reddit User Browsing Model}
\label{app:reddit-browsing-model}
The size and structure of the conversation graphs present challenges for representing prior context. Naively summarizing all preceding comments is both computationally expensive and likely to include content irrelevant to the comment at hand, due to branching sub-conversations. 

Based on Reddit's hierarchical UI, we assume that when readers encounter a storytelling comment, they have viewed the conversation's initial post, the ancestral chain of the comment’s parent, and the parent’s chronologically preceding peer comments. For efficiency, we approximate this reader model by adopting an iterative summarization strategy with GPT-4o. We first summarize the initial post. For each storytelling comment, we then summarize up to $k=5$ ancestral parent comments and up to $k=5$ prior peer comments, integrating these with the initial post summary into a single conversational context summary.

\subsection{Context Summarization Validation}
\label{app:ctx-sum-val}
\subsubsection{Evaluation Criteria}
\label{app:summarization-criteria}
Both criteria below use Likert rating from 1 to 5 (with 1 being worst and 5 being best)

\paragraph{Consistency}
``the factual alignment between the summary and the summarized source. A factually consistent summary contains only statements that are entailed by the source document.''

\paragraph{Relevance}
``selection of important content from the source. The summary should include only important information from the source document.''
If there is more important information than can reasonably fit within our short summaries, then the relevance score should reward summaries for selecting the most important information. 

\subsubsection{Ratings}
\begin{figure*}[t]
    \centering
    \includegraphics[width=\linewidth]{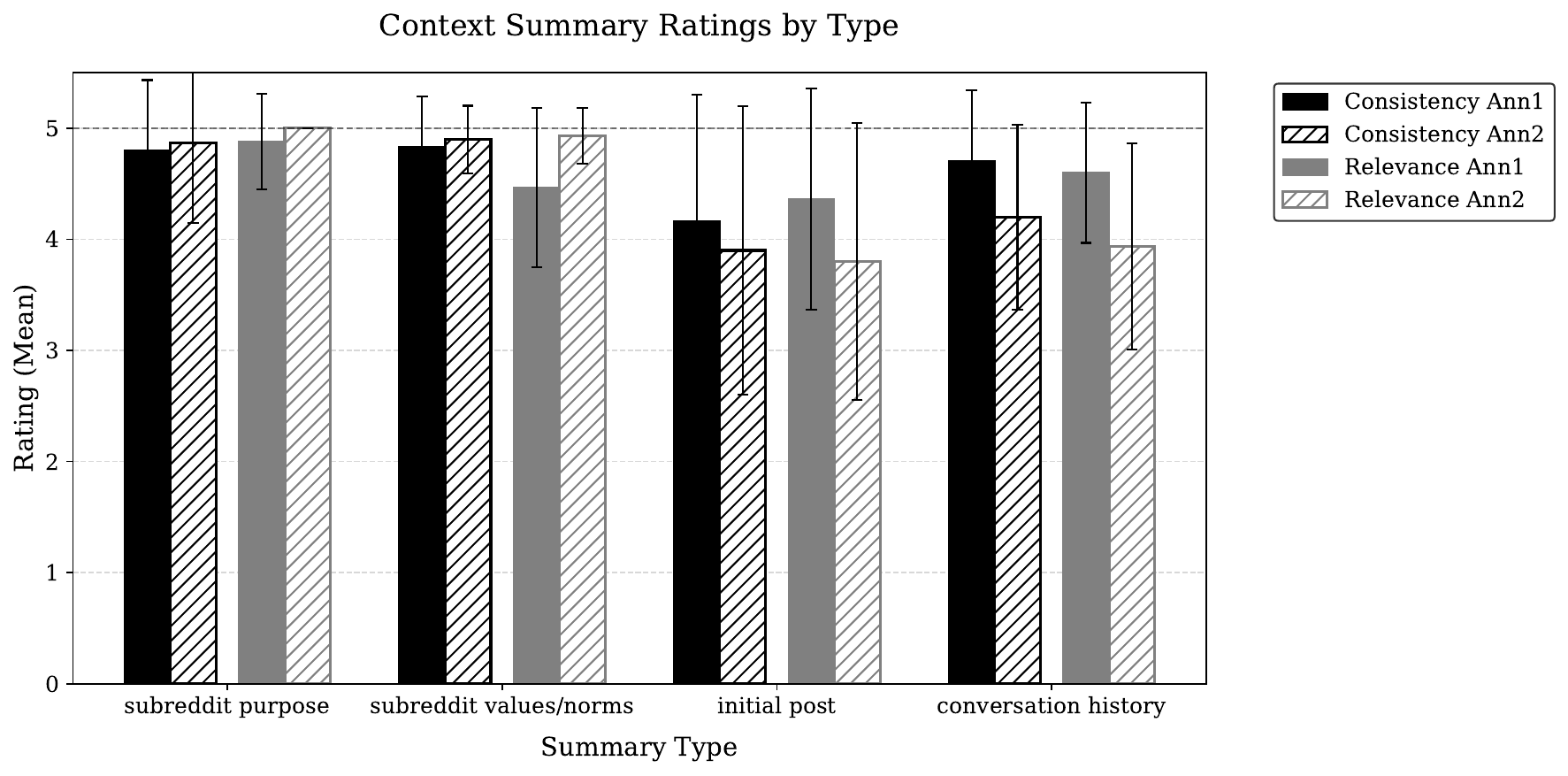}
    \caption{Mean ``Consistency'' and ``Relevance'' context summary ratings for annotator1 ($N=50$) and annotator2 ($N=30$).}
    \label{fig:ctx_sum_ann_raw}
\end{figure*}
Fig.~\ref{fig:ctx_sum_ann_raw} shows the results of the context summary annotation. Two raters, annotator1 ($N=50$) and annotator2 ($N=30$), independently evaluated each summary for consistency and relevance.

\subsubsection{Inter-Annotator Agreement}
\begin{table*}[htbp]
\centering
\small
\begin{tabular}{lrrrrr rrrrr}
\toprule
\multirow{2}{*}{Summary Type} & \multicolumn{5}{c}{Consistency} & \multicolumn{5}{c}{Relevance} \\
\cmidrule(lr){2-6} \cmidrule(lr){7-11}
& \%Agr & \%±1 & $\kappa_b$ & AC$_2$ & $\kappa_c$ & \%Agr & \%±1 & $\kappa_b$ & AC$_2$ & $\kappa_c$ \\
\midrule
Subreddit Purpose & 90.0 & 100.0 & 0.95 & 0.98 & 0.79 & 90.0 & 96.7 & 0.85 & 0.94 & 0.00 \\
Subreddit Norms/Values & 75.9 & 96.6 & 0.72 & 0.87 & -0.12 & 55.2 & 93.1 & 0.63 & 0.82 & -0.01 \\
Initial Post & 50.0 & 93.3 & 0.77 & 0.84 & 0.77 & 46.7 & 80.0 & 0.60 & 0.71 & 0.50 \\
Conversation History & 56.7 & 86.7 & 0.63 & 0.77 & 0.41 & 30.0 & 86.7 & 0.43 & 0.58 & 0.25 \\
\bottomrule
\end{tabular}
\caption{Inter-rater Agreement Metrics by Summary Type and Dimension. \%Agr = exact agreement; \%±1 = agreement within 1 point; $\kappa_b$ = Brennan-Prediger (ordinal); AC$_2$ = Gwet's AC$_2$ (ordinal); $\kappa_c$ = Cohen's kappa (quadratic).}
\label{tab:context_summarization_iaa}
\end{table*}

We report standard inter-annotator agreement (IAA) metrics in Table \ref{tab:context_summarization_iaa}. Percent agreement was high, but a quadratic-weighted Cohen’s $\kappa$ \cite{Cohen1968-ja} was low due to the ``kappa paradox'' \cite{Feinstein1990-bf}. This paradox occurs when most ratings fall in the same category—here, the maximum score of 5—so chance agreement is overestimated and kappa is artificially deflated. To address this, we also report the Brennan–Prediger coefficient, $\kappa_b$ \cite{Brennan1981-mc}, and Gwet’s $AC_2$ \cite{Gwet2008-dp, Gwet2014-wz}, which still correct for chance agreement but rely on less punitive assumptions than Cohen’s $\kappa$. Taken together, percent agreement and these alternative chance-corrected coefficients indicate moderately high IAA.

\subsubsection{Error Analysis}
Below, we qualitatively analyze common errors observed in the community/conversation context summaries during human validation.

\paragraph{\textsc{hallucination\_for\_short\_texts}}
When source texts are extremely short, the GPT-4o sometimes fabricates plausible but unsupported details to produce a summary-like output. For example, the post “just a heads up for fans of the show'' was summarized as ``The post informs fans of a show airing at a new time in the UK,'' adding geographic detail without basis. This reflects a tendency to overgenerate when there is minimal input context. We note that this may have been inadvertently exacerbated by our summarization prompt, which instructs models never to refuse and always provide a summary. 

\paragraph{\textsc{underemphasizing\_disagreement}}
This error occurs when the summary correctly identifies the existence of debate or controversy in prior conversation, but does not provide sufficiently precise information about the contour of the debate, and in particular, what stance was expressed by the last speaker.

In one example, a conversation involves a debate around which MMA fighter should have won the first round, and the last speaker asserted that ``Conor should have won the round.'' However, the generated conversation summary elides this salient information: ``The conversation so far revolves around the Conor McGregor vs. Chad Mendes fight, discussing Mendes' underrated skills, McGregor's strong performance, and the effects of a last-minute opponent change. Users debate betting outcomes and who won the first round, emphasizing the fight's competitive nature.''

\paragraph{\textsc{poor\_value\_abstraction}}
Many subreddits’ community guidelines contain long lists of rules. Our GPT-4o–based value/norm summarization tends to extract only a few key rules or norms, rather than abstracting them into higher-level statements about underlying values. Additionally, because many subreddits share basic rules, summaries across different subreddits can appear quite similar. Consequently, our summaries are not always effective at differentiating subreddits in terms of values. While this is more a limitation than a failure, it means that certain value/norm summaries are less informative than desired.

\section{Model Training and Inference Details}
\label{app-training-details}
We followed a similar supervised finetuning (SFT) setup to train both \ssfGenerator and \ssfClassifier. We first describe shared hyperparameters, followed by model-specific details in the subsections below.

We conducted LoRA finetuning (rank=$16$, alpha=$16$, dropout=$0.05$) on Llama-3.1-8B-Instruct,\footnote{\texttt{meta-llama/Meta-Llama-3.1-8B-Instruct}} targeting the following modules: \texttt{q\_proj}, \texttt{k\_proj}, \texttt{v\_proj}, \texttt{o\_proj}, \texttt{gate\_proj}, \texttt{down\_proj}, and \texttt{up\_proj}.
We used a batch size of $8$, a warmup ratio of $0.1$, and a learning rate of $5e-5$.

\subsection{SSF-Generator Training}
\label{app-ssf-generator-training-details}
Since each story included up to three GPT-4o-generated reference instances, we trained \ssfGenerator for a single epoch with a sequence cutoff length of $2048$ tokens.

\subsection{SSF-Classifier Training}
\label{app-ssf-classifier-training-details}
We trained \ssfClassifier for $3$ epochs using a sequence cutoff length of $1536$ tokens.

\subsection{Inference}
At inference time, we use a decoding temperature of $0$ and \texttt{top\_p} $= 1$.

\section{Inference Generation Details}
\subsection{Inference Validation: Prolific Details}
\label{app:prolific-annotation-task}

\subsubsection{Recruitment Materials}
\definecolor{lightblue}{RGB}{235, 245, 255}
\begin{tcolorbox}[colback=lightblue,colframe=RoyalBlue!75!white,title=Prolific Study Advertisement,breakable]
\small
Welcome! This is a research study about storytelling on social media. \\

We will show you a text from a social media conversation that contains storytelling. We will also describe the conversational context in which the text was posted or commented. \\

Your task will be to reason about (1) why the author might have posted/commented the story or (2) how a reader might plausibly react to the story. \\

The purpose of this study is to better understand the social functions of storytelling in online conversations. Your (anonymous) annotations will be included in a dataset released for other researchers. We will use this dataset to study stories computationally. This research study is conducted by Joel Mire at Carnegie Mellon University. \\

If you participate, you will be expected to: \\
- Read social media texts containing stories + conversational context summaries \\
- Rate the plausibility of candidate answers to questions about narrative intents or reader reactions. \\

Please do not use AI tools like ChatGPT to answer these questions because that could mess up our scientific results. We’re interested in your answers, not a bot’s. We really appreciate your work!
\normalsize
\end{tcolorbox}

\subsubsection{Consent Form}
\begin{tcolorbox}[colback=lightblue,colframe=RoyalBlue!75!white,title=Consent Form,breakable]
\small
This task is part of a research study (STUDY2025\_00000026) conducted by Joel Mire at Carnegie Mellon University and funded by Dr. Maarten Sap. \\

Summary \\
Welcome! This is a study about storytelling in social media. We will show you a text from a social media conversation that contains storytelling and describe the context in which it was posted or commented. Your task is to reason about (1) why the author posted/commented the story or (2) how a reader might plausibly react. Your (anonymous) answers will be included in a dataset for research purposes. Please do not use AI tools to answer, as we want your answers, not synthetic ones. Thank you! \\

Purpose \\
This research aims to better understand the social functions of storytelling on social media. It is unclear how storytelling varies across communities, and studying it computationally at a large scale is challenging. Our research seeks to address these gaps. \\

Procedures \\
If you participate, you will: \\
• Read categories/definitions for narrative intent and reader reaction. \\
• Analyze social media texts with context summaries. \\
• Answer questions about plausible intents or reader reactions. \\
• Rate plausibility of answers to these questions. \\
The study is expected to take 7 minutes. \\

Participant Requirements \\
You must be aged 18+, read English proficiently, and reside in the U.S. You must also be in the U.S. while completing the survey on Prolific. \\

Risks \\
The risks of participation are minimal, on par with those of daily life or other online activities. However, some content from online forums may be upsetting or NSFW despite filters. \\

Benefits \\
There may be no personal benefits, but the research may contribute to scientific understanding of storytelling on social media. \\

Compensation \& Costs \\
Compensation is $1.75$, based on a $\$15$/hour rate, for the estimated 7-minute study. Payment will be made through the Prolific platform. There is no cost to participate. \\

Future Use of Information \\
Your data may be used in future studies or shared with other researchers in a manner that does not identify you. Your annotations will be used to evaluate computational models and included in publicly released datasets. \\

Confidentiality \\
No personally identifiable information will be collected. However, Prolific, which is not affiliated with CMU, may collect your identifiable information as per its terms of use. \\

Researchers Involved \\
• Joel Mire, Carnegie Mellon University \\
• Maarten Sap, Carnegie Mellon University \\
• Andrew Piper, McGill University \\
• Maria Antoniak, University of Copenhagen \\
• Steve Wilson, University of Michigan – Flint \\
• Zexin Ma, University of Connecticut \\
• Achyut Ganti, Oakland University \\

Right to Ask Questions \& Contact Information \\
If you have questions or wish to withdraw participation, contact us via the Prolific platform or email Joel Mire at jmire@andrew.cmu.edu. For questions about your rights as a participant, contact the Office of Research Integrity and Compliance at Carnegie Mellon University. \\

Voluntary Participation \\
Your participation is voluntary, and you may discontinue at any time. You may print a copy of this consent form for your records. \\

I am age 18 or older. \\
Yes or No \\

I have read and understand the information above. \\
Yes or No \\

I want to participate in this research and continue with the study. \\
Yes or No
\normalsize
\end{tcolorbox}

\subsubsection{Example: Annotation Instance}
Here, we provide screenshots from the annotation task(s), including the plausibility rating and human-written variants of the task.

\begin{figure*}[t]
    \centering
    \includegraphics[width=\textwidth]{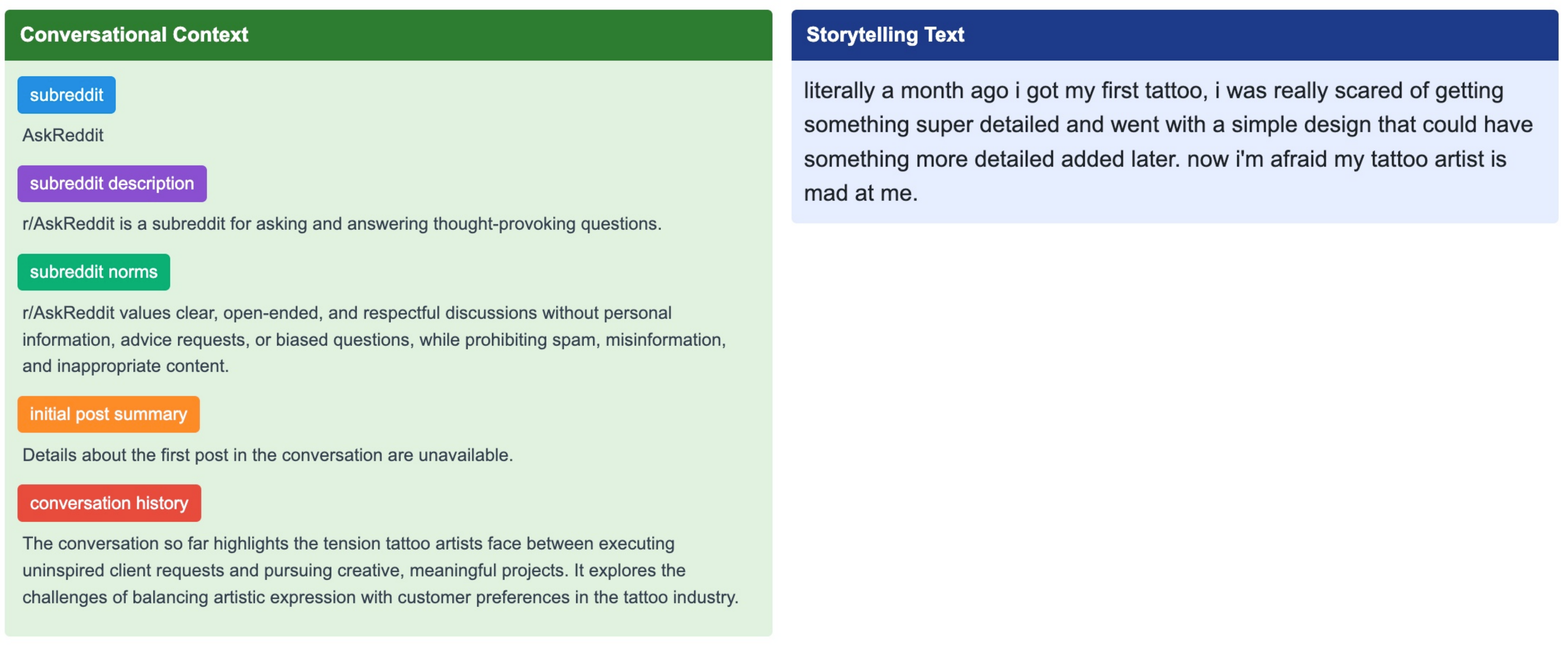}
    \caption{An example of a story, plus the available community and conversational context, presented to an annotator at the start of either annotation task}
\label{fig:example-context_and_story}
\end{figure*}

In either case, the task starts with the user reading a story in its community and conversational context (Fig. \ref{fig:example-context_and_story}).

\subsubsection{Example: Plausibility Rating}
\begin{figure*}[h]
    \centering
    \includegraphics[width=0.8\textwidth]{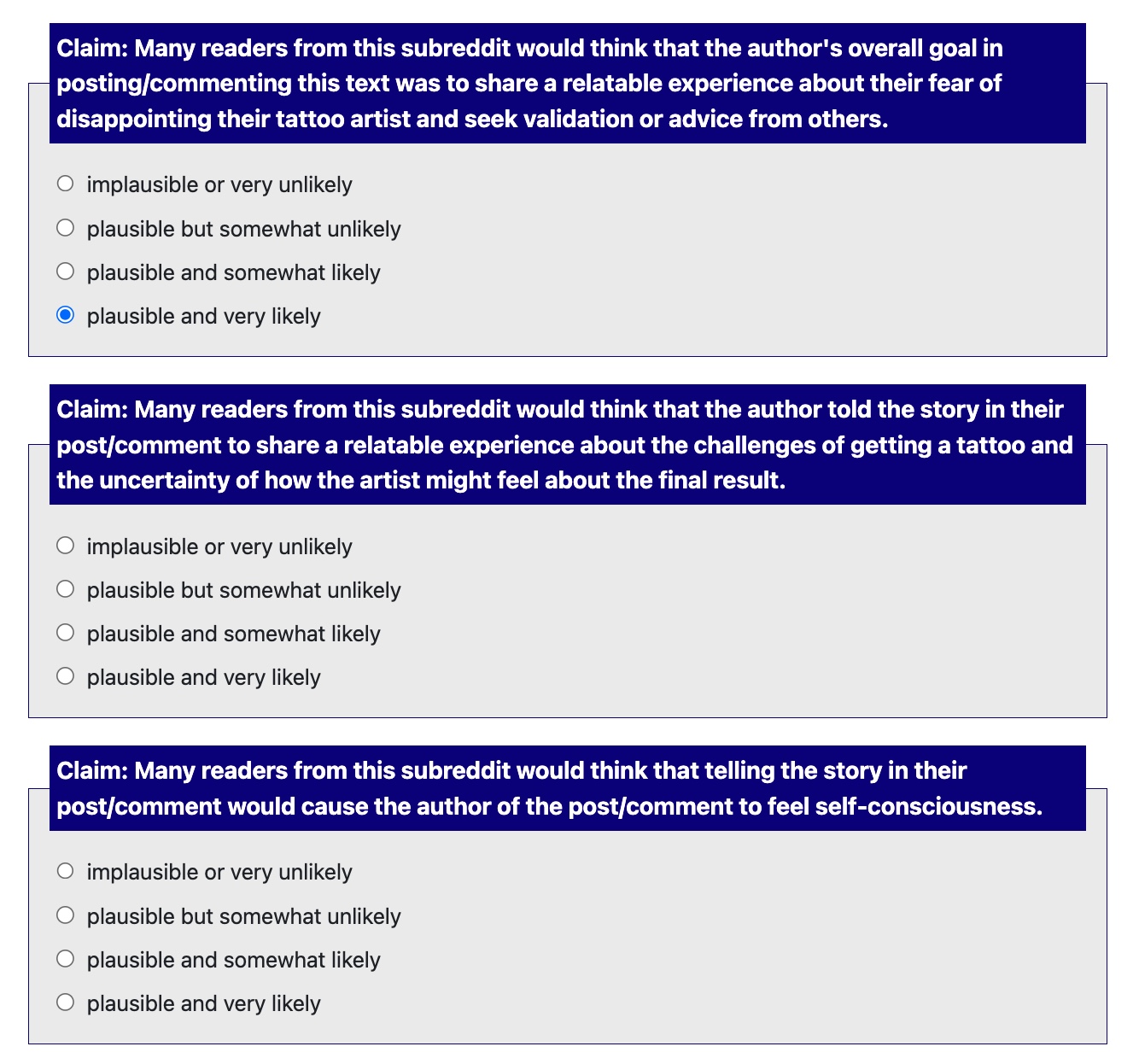}
    \caption{Examples of plausibility rating questions for the overall\_goal, narrative\_intent, and author\_emotional\_response \ssfTaxonomy dimensions.}
\label{fig:example-plausibility-rating}
\end{figure*}

Fig. \ref{fig:example-plausibility-rating} shows the format of the dimension-specific plausibility rating questions.

\subsubsection{Example: Human-Written Task}
\begin{figure*}[t]
    \centering
    \includegraphics[width=\textwidth]{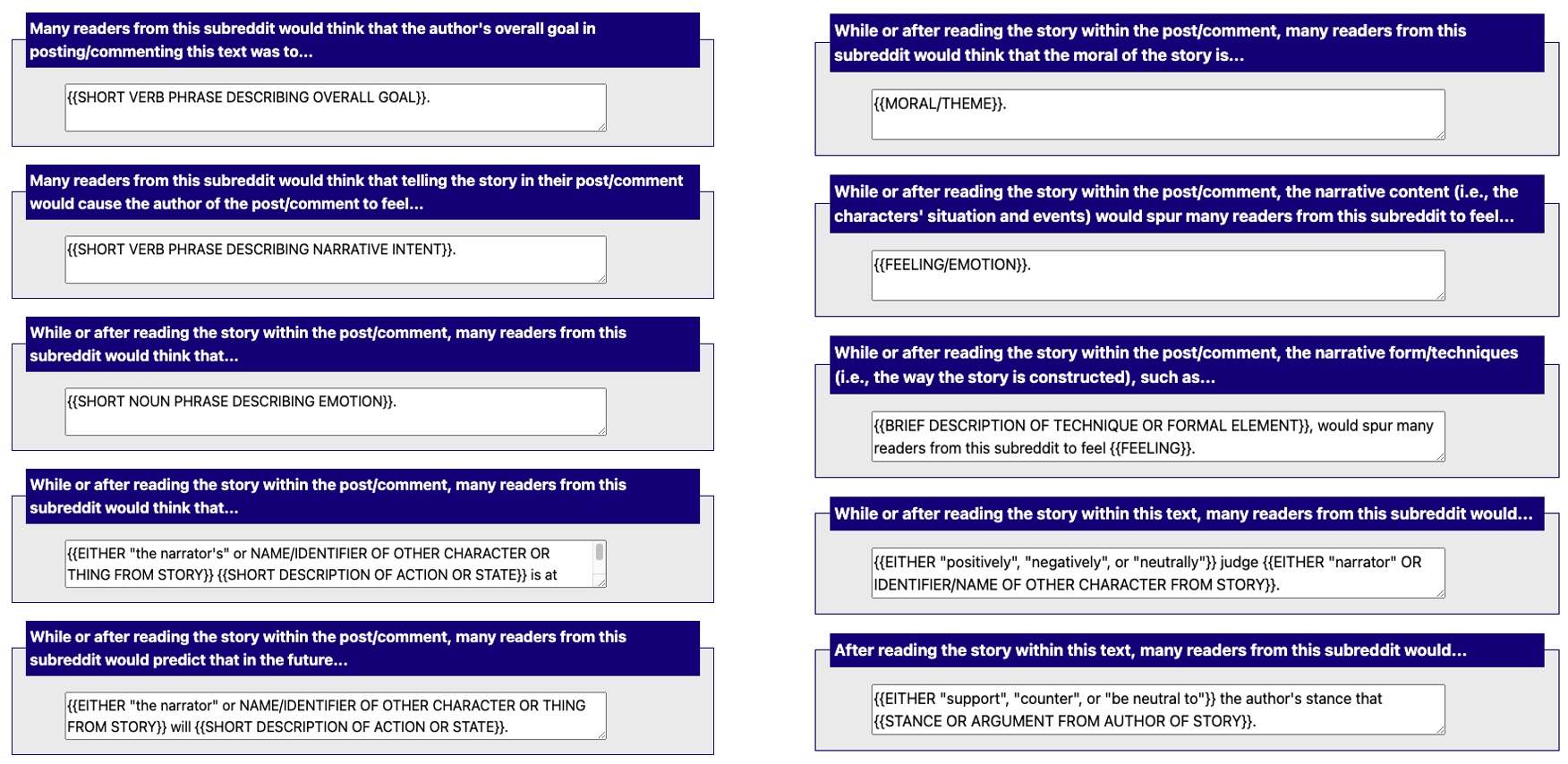}
    \caption{An example of the dimension-specific questions asking annotators to substitute the placeholder for a free-text inference.}
\label{fig:example-human-written}
\end{figure*}

Fig. \ref{fig:example-human-written} shows the format of the dimension-specific human-written tasks.

\subsubsection{Participant Demographics}
% \begin{table*}[t]
% \centering
% \small
% \caption{Demographic Characteristics Across Studies}
% \begin{tabular}{|l|c|c|c|}
% \hline
% Demographic & Human Written (N=39) & GPT-4o Ratings (N=319) & \ssfGenerator Ratings (N=110) \\
% \hline
% \textbf{Gender} & & & \\
% Female & 51.3\% & 50.2\% & 50.0\% \\
% Male & 48.7\% & 49.8\% & 50.0\% \\
% \hline
% \textbf{Age} & & & \\
% 18-24 & 5.1\% & 11.9\% & 4.5\% \\
% 25-34 & 25.6\% & 17.9\% & 20.9\% \\
% 35-44 & 38.5\% & 16.6\% & 30.0\% \\
% 45-54 & 17.9\% & 15.7\% & 25.5\% \\
% 55-64 & 10.3\% & 27.9\% & 10.9\% \\
% 65-74 & 0.0\% & 9.1\% & 7.3\% \\
% 75+ & 2.6\% & 0.9\% & 0.9\% \\
% \hline
% \textbf{Race/Ethnicity} & & & \\
% White & 79.5\% & 63.3\% & 71.8\% \\
% Black & 10.3\% & 11.6\% & 14.5\% \\
% Mixed & 2.6\% & 11.0\% & 2.7\% \\
% Asian & 5.1\% & 6.6\% & 8.2\% \\
% Other & 0.0\% & 7.5\% & 0.9\% \\
% DATA\_EXPIRED & 2.6\% & 0.0\% & 1.8\% \\
% \hline
% \end{tabular}
% \caption{Demographic information from the Prolific surveys. [left] small-scale human-written inferences, [middle] large-scale representative U.S. sample for ratings of GPT-4o inferences; [right] medium-scale sample for ratings of \ssfGenerator inferences. Sample sizes reflect full annotator pools for each study before the known-implausible filtering described in Section \ref{sec:ssf-modeling}.}
% \label{tab:demographics}
% \end{table*}

% \usepackage{booktabs} % in preamble

\begin{table*}[t]
\centering
\small
% \caption{Demographic Characteristics Across Studies}

\begin{tabular}{lrrr}
\toprule
Demographic & Human Written (N=39) & GPT-4o Ratings (N=319) & \ssfGenerator Ratings (N=110) \\
\midrule

\textbf{Gender} & & & \\
Female & 51.3\% & 50.2\% & 50.0\% \\
Male & 48.7\% & 49.8\% & 50.0\% \\

\midrule
\textbf{Age} & & & \\
18-24 & 5.1\% & 11.9\% & 4.5\% \\
25-34 & 25.6\% & 17.9\% & 20.9\% \\
35-44 & 38.5\% & 16.6\% & 30.0\% \\
45-54 & 17.9\% & 15.7\% & 25.5\% \\
55-64 & 10.3\% & 27.9\% & 10.9\% \\
65-74 & 0.0\% & 9.1\% & 7.3\% \\
75+ & 2.6\% & 0.9\% & 0.9\% \\

\midrule
\textbf{Race/Ethnicity} & & & \\
White & 79.5\% & 63.3\% & 71.8\% \\
Black & 10.3\% & 11.6\% & 14.5\% \\
Mixed & 2.6\% & 11.0\% & 2.7\% \\
Asian & 5.1\% & 6.6\% & 8.2\% \\
Other & 0.0\% & 7.5\% & 0.9\% \\
DATA\_EXPIRED & 2.6\% & 0.0\% & 1.8\% \\

\bottomrule
\end{tabular}

\caption{Demographic information from the Prolific surveys. [left] small-scale human-written inferences, [middle] large-scale representative U.S. sample for ratings of GPT-4o inferences; [right] medium-scale sample for ratings of \ssfGenerator inferences. Sample sizes reflect full annotator pools for each study before the known-implausible filtering described in Section \ref{sec:ssf-modeling}.}

\label{tab:demographics}
\end{table*}

Lastly, we present the demographics for our three Prolific surveys in Table \ref{tab:demographics}.

\subsubsection{Quality Filtering with Known-Implausible Inferences}
\label{app:prolific-quality-filter}
Randomly, we inject one known-implausible inference for $1/10$ taxonomy dimensions. If the annotator marks the known-implausible inference as ``somewhat likely'' or ``very likely'', we conclude that they have not paid attention and discard all $10$ of their annotations for the given story. Through this filtering mechanism, we mitigate the risk of cognitive biases affecting our results, such as confirmation bias (if an annotator believes that most are plausible, they may search for interpretations to justify the belief), automation bias (if they suspect that the inferences were machine generated---even though this was not specified---and have a bias toward trusting automated systems), and any annotators who simply clicked through the task without reading the inferences closely.

See App. \ref{app:prompt-templates-inference-generation-implausible} for the prompt template for the known implausible generations with GPT-4o.

\subsubsection{Human-Written Inferences vs. GPT-4o Inferences}
\begin{table*}[h]
\small
\centering
\begin{tabular}{lrrrrr}
\toprule
Dimension & Cosine Sim. & BERTScore & BLEU & METEOR & N \\
\midrule
overall\_goal & 0.403 & 0.862 & 0.095 & 0.147 & 54 \\
narrative\_intent & 0.368 & 0.862 & 0.097 & 0.127 & 54 \\
author\_emotional\_response & 0.300 & 0.861 & 0.231 & 0.070 & 54 \\
character\_appraisal & 0.568 & 0.889 & 0.480 & 0.210 & 48 \\
causal\_explanation & 0.394 & 0.866 & 0.118 & 0.170 & 41 \\
prediction & 0.460 & 0.878 & 0.335 & 0.244 & 45 \\
stance & 0.540 & 0.903 & 0.381 & 0.232 & 45 \\
moral & 0.388 & 0.855 & 0.071 & 0.113 & 54 \\
narrative\_feeling & 0.313 & 0.839 & 0.067 & 0.089 & 54 \\
aesthetic\_feeling & 0.252 & 0.847 & 0.130 & 0.081 & 45 \\
\midrule
Overall & 0.399 & 0.866 & 0.201 & 0.148 & - \\
\bottomrule
\end{tabular}
\caption{Comparison of human-written inferences to GPT-4o generated inferences. We report cosine similarity, BERTScore \cite{Zhang2019-tb}, $BLEU_3$ \cite{Papineni2002-hn}, and METEOR \cite{Banerjee2005METEORAA}.}
\label{tab:gpt4o_x_prolific_hw}
\end{table*}
We also collect a small number of human-written inferences and compare them to the GPT-4o inferences (Table \ref{tab:gpt4o_x_prolific_hw}).

\subsection{Error Analysis}
\label{app:ssf-gen-err}
We inspect the $100$ \ssfGenerator inferences with the lowest average plausibility ratings from Prolific human annotators to identify common failure modes. 

\paragraph{\textsc{overemphasize\_reader\_investment}} occurs when the reader is assumed to be exceedingly emotionally invested in a relatively banal story, or assumed to automatically mirror the emotions of the author. For example, in one analytical discussion about a soccer player's performance, a commenter recounted an unsuccessful play, clearly attributing fault to the player. \ssfGenerator's inference claimed that many readers would feel ``frustrated'' in response to the player's performance. However, given the analytical tone and the uncertain assessment of the player, presuming such strong reader emotions is likely an overreach.

\paragraph{\textsc{overemphasize\_impact}} means inferring that either the content or the act of telling a particular story will galvanize either the author or readers to change their beliefs or take specific actions.

A typical example is assuming the author will \textit{continue} a past behavior. For instance, in one conversation, an author told a story about their gecko. \ssfGenerator inferred that the author was likely to share another heartwarming pet story in the future. While this is possible, it is not necessarily probable, as it makes several assumptions about the author's future behavior that are not foreshadowed by their prior comment---there is no promise of a future story or hint that there are other stories the author wants to tell.

In another discussion about video games, a commenter mentioned owning some games but not having played them. The inference suggested that the conversation would prompt the commenter to start playing, which is arguably an unwarranted assumption about the conversation’s impact.

\paragraph{\textsc{assume\_emotional/moral\_closure}} refers to inferences that preempt or assume the necessity of emotional or moral resolution to a situation that either remains unresolved or is not significant enough to require closure.

For example, someone told a story about a recent issue with loss of feeling in their fingers that inhibited their guitar practice. Although the status of the medical issue and the commenter's emotional state were unclear, the \ssfGenerator inferred that the moral of the story relates to the role of community support in alleviating feelings of isolation and providing emotional relief. This may be true, but it is a jump to assume the commenter shared this perspective.

\paragraph{\textsc{assume\_different\_consensus}} occurs when the model assumes there is consensus (or at least a majority perspective) when an issue is actually quite contentious among readers.

For example, in a story where someone drilled holes in a park slide to prevent water pooling, readers appeared to disagree on whether the driller's actions were justified and whether the town’s response (temporarily closing the park and spending money on a replacement slide) was reasonable. \ssfGenerator's inferences, however, held to the idea that the town overreacted. This ostensibly led some readers who believed the town responded appropriately to rate \ssfGenerator's inferences as unlikely.

\paragraph{\textsc{ignore\_context}} refers to cases in which an inference seems plausible in isolation but becomes implausible in light of the broader conversational context.

For example, a story described an older parent running an errand to pick up a beverage to share with their partner outdoors. The inference remarked on the peacefulness of the moment, which seems reasonable on its own. However, earlier conversation revealed concerns about high-risk behaviors in people with dementia and the difficulties in preventing them from driving, making the actual context far more anxiety-inducing than the inference suggests.

\paragraph{\textsc{wrong\_focalization/attribution}}

In one story, a person was complaining about their bad experience with a taxi driver in a foreign country. While the story was mostly meant to vent and emphasize the challenges of travel from the perspective of the commenter, the inference instead suggested that the theme was that professionalism and politeness are important. While this is connected to the story, it misreads the tone/focalization of the post, which is about the challenges of traveling, not normative commentary about professionalism. 

\paragraph{\textsc{no\_obvious\_issue}}
Many inferences do not have obvious problems, and yet readers rated them as implausible or highly unlikely. This indicates inherent subjectivity in the plausibility rating task. Perceptions of plausibility are not universal, highlighting fundamental tensions in our approach that represent a compromise between two extremes: universalist and individualist approaches to modeling reader response. 

\begin{table*}[h]
\centering
\small
\begin{tabular}{lrrrrrr}
\toprule
comparison & $n$ & $\bar{x}_1$ & $\bar{x}_2$ & $t$ & $d$ & $p_{\text{holm}}$ \\
\midrule
    comm\_ctx vs no\_ctx & 2865 & 0.521 & 0.517 & 2.251 & 0.030 & \textbf{0.024} \\
    conv\_ctx vs no\_ctx & 2943 & 0.539 & 0.517 & 11.068 & 0.159 & \textbf{1.27e-27} \\
    comm+conv\_ctx vs comm\_ctx & 2877 & 0.542 & 0.521 & 10.269 & 0.151 & \textbf{3.83e-24} \\
    comm+conv\_ctx vs conv\_ctx & 2959 & 0.542 & 0.539 & 1.522 & 0.019 & 0.064 \\
    comm+conv\_ctx vs no\_ctx & 2947 & 0.542 & 0.517 & 12.078 & 0.178 & \textbf{2.03e-32} \\
\bottomrule
\end{tabular}
\caption{\ssfGenerator SFT Distillation Context Ablation Results. We compare \ssfGenerator variants' inferences against GPT-4o reference inferences via one-sided paired t-tests of cosine similarities, based on the hypothesis that more context will lead to greater inference similarity than less context. Statistically significant results (Holm-Bonferroni corrected) are in \textbf{bold}.}
\label{tab:ablation_ttests}
\end{table*}

\subsection{SFT Distillation Context Ablations}
\label{app:ctx-ablations}

To test whether incorporating community and conversational context leads to more successful SFT distillation of \ssfGenerator, we ablate context and finetune on the same GPT-4o-generated reference inferences that were extensively validated in our human studies. We compute cosine similarity (using Sentence-BERT \texttt{all-MiniLM-L6-v2} embeddings) between each \ssfGenerator variant’s inferences and the reference inferences. We then run one-sided t-tests under the hypothesis that variants with more context will show greater cosine similarity with the references. We apply a Holm–Bonferroni correction \cite{Holm1979-xk} across test conditions and report results on the \ssfSplitCorpus test split in Table \ref{tab:ablation_ttests}.

Partial context yields significantly higher similarity to the reference inferences than no context, with a stronger effect for conversational context. Comparing full context to partial-context and no-context settings, full context produces significantly more similar inferences than both the community-context and no-context conditions. Taken together, these results suggest that modeling social context is important for generating plausible inferences aligned with human perceptions, and that conversational history is particularly valuable.

\section{Inference Classification Details}
\subsection{K-Shot Prompting Strategy for Inference Classification}
\label{app:k-shot-prompting-details}
We adopt a mixed-method sampling strategy. First, we randomly sample $k/4$ human-annotated demonstrations from the validation split of \ssfSplitCorpus. Second, we sample $k/2$ demonstrations via the maximum marginal relevance (MMR) criterion \cite{Carbonell2017-oo}. The MMR score is similar to a pure semantic similarity score, except it adds a penalty for sampling a demonstration that is similar to a previously selected demonstration. In this way, it samples semantically similar yet diverse results. Formally, MMR is defined as:

\begin{equation}
\label{eq:mmr}
\begin{aligned}
MMR = \arg\max_{D_i \in R \setminus S} \, 
   \lambda \, \text{Sim}(D_i, Q) \\
   - (1 - \lambda) \max_{D_j \in S} \text{Sim}(D_i, D_j)
\end{aligned}
\end{equation}

where $R$ is the full set of candidate examples, $S$ is the set of selected candidates, $D_i$ is the current candidate, $Q$ is the current inference to classify, and $Sim$ is cosine similarity.

Finally, we sample the remaining $k/4$ demonstrations based purely on semantic similarity, defined as the cosine similarity between text embeddings generated by Sentence-BERT \texttt{all-MiniLM-L6-v2}.

We vary $k$ across \ssfTaxonomy dimensions because classification difficulty and label-set size differ from one dimension to another. See  Table \ref{tab:demos_by_dimension} for details.

\begin{table}[h]
\small
\centering
\begin{tabular}{lr}
\toprule
\textbf{Dimension} & \textbf{k} \\
\midrule
overall\_goal & 30 \\
narrative\_intent & 30 \\
author\_emotional\_response & 20 \\
character\_appraisal & 10 \\
causal\_explanation & 30 \\
prediction & 15 \\
stance & 10 \\
moral & 30 \\
narrative\_feeling & 20 \\
aesthetic\_feeling & 15 \\
\bottomrule
\end{tabular}
\caption{Number of demonstrations ($k$) used for each \ssfTaxonomy dimension for the GPT-4.1 $k$-shot reference inference classification.}
\label{tab:demos_by_dimension}
\end{table}

\subsection{Inference Classification Annotation Guidelines}
\label{app:tax-ann-guide}
In this multi-label annotation task, we map templated inferences (generated by an LLM or human in response to a story) onto the sublabels in \ssfTaxonomy.

\definecolor{darkolivegreen}{RGB}{85,107,47}
\definecolor{lightolivegreen}{RGB}{245,250,240}

\subsubsection{\colorbox{goal-color}{Overall Goal}}
\begin{tcolorbox}[colback=brown!5!white,colframe=brown!50!black,title=Taxonomy Classification Annotation Guide,breakable]
\small
• sharing an experience to ``highlight'' a normative or controversial perspective → ``persuade\_debate'' \\
• sharing a perspective/advice in a casual/helpful manner, without the sense that the narrator is personally invested in getting readers to take their advice  →  ``provide\_info\_support'' \\
• sharing an experience to express, explain, or justify one’s identity, core beliefs, values, or emotions; or highlight why one feels a certain way; nostalgia → ``affirm\_identity\_self'' \\
• trying to ``connect'' or to ``relate'' or be ``relatable'' to other people; apologizing → both ``provide\_emotional\_support'' and ``request\_emotional\_support'' \\
• if instead of relating to people, the focus in on sharing a ``related'' (i.e. topically relevant) story, just label ``provide\_experiential\_support'' \\
• trying to ``empathize with'' → ``provide\_emotional\_support'' \\
• references to humor; lighthearted or memorable stories → ``entertain'' \\
• if expressing disappointment and it is not obvious that they are just venting/expressing themself without concern for how people respond → implicit ``request\_emotional\_support'' \\
• sharing/highlighting a personal perspective doesn’t necessarily imply ``provide\_experiential\_accounts''. \\
\normalsize
\end{tcolorbox}

\subsubsection{\colorbox{intent-color}{Narrative Intent}}
\begin{tcolorbox}[colback=brown!5!white,colframe=brown!50!black,title=Taxonomy Classification Annotation Guide,breakable]
\small
• explaining events, providing context, updates, or describing what happened **to correct misunderstandings or fill information gaps** (informational intent) → ``clarify\_what\_transpired'' \\
\hspace*{1.5em} • **do not use ``clarify\_what\_transpired'' simply because events are referenced; only when the primary intent is corrective or to add new facts/info or a personal account to the topic** \\
• revealing core values, identifications, or moments of personal growth or self-awareness → ``show\_identity'' \\
• drawing explicit broader conclusions or offering advice/lessons that others could apply beyond the specific situation described → ``justify\_challenge\_offer\_belief\_norm'' \\
• advocating for an idea, belief, or opinion (e.g., by providing evidence to defend a new or existing claim); defending/supporting/explaining one's interpretation; contradicting misconceptions, challenging established narratives, or disputing commonly held beliefs (persuasive/argumentative intent) → ``justify\_challenge\_offer\_belief\_norm'' \\
• venting or intentionally expressing an intense emotion (e.g., anger, sadness, relief, etc) → ``release\_pent\_up\_emotions'' \\
• humor, lighthearted → ``entertain'' \\
• seeking reassurance, validation, or emotional support (often through sharing struggles, asking for advice, expressing uncertainty, describing awkward/difficult situations, or revealing vulnerability) → ``convey\_emotional\_support\_need'' \\
\hspace*{1.5em} • ``release\_pent\_up\_emotions'' is for cathartic expression; ``convey\_emotional\_support\_need'' is for seeking comfort/help \\
\hspace*{1.5em} • merely seeking informational advice does NOT qualify as ``convey\_emotional\_support\_need'' \\
• creating connection by sharing relatable experiences to bond with others'' → ``convey\_similar\_experience'' if there is a clear signal in the response suggesting this intent \\
\hspace*{1.5em} • do not assume ``convey\_similar\_experience'' based on outside knowledge \\
• seeking informational (as opposed to emotional) support/advice, speculation → out of scope, so return empty list if no other labels apply \\
• **focus on prominent intent(s); secondary purposes should only be included if relatively substantial** \\
• when correcting misinformation or explaining events, consider whether the primary intent is neutral explanation (clarify\_what\_transpired) or taking a stance to persuade/challenge (justify\_challenge\_offer\_belief\_norm). Both can apply when someone explains facts AND argues a position.
\normalsize
\end{tcolorbox}

\subsubsection{\colorbox{emotion-color}{Author Emotional Response} and \colorbox{feeling-color}{Narrative Feeling}}
\begin{tcolorbox}[colback=brown!5!white,colframe=brown!50!black,title=Taxonomy Classification Annotation Guide,breakable]
\small
• amusement → ``joy'', ``appreciation'' \\
• validation and agreement → ``relief'', ``pride'', ``connection'' \\
• nostalgia → ``sadness'', ``joy'', ``appreciation'', ``connection'' \\
• satisfaction → ``pride'' (could sometimes also point to ``joy'', ``relief'', and/or ``appreciation'' depending on context) \\
• hopelessness → ``fear'', ``sadness'' \\
• self-consciousness about one’s appearance → ``fear'', ``disgust'' \\
• frustration or exasperation → ``anger'' \\
• if the anger/exasperation/disbelief is visceral or especially intense AND directed toward a misbehaving third party whose behavior is explicitly/implicitly considered very offensive → also add ``disgust'' \\
• concerned, anxious, defensive → ``fear'' \\
• cautious → usually ``fear'' but sometimes *nothing* (e.g., ``cautious skepticism'' is nothing) \\
• regret → ``sadness'' \\
• conflicted, embarrassed → depends on context, but often ``guilt'' \\
• confidence → usually ``hope'' but could be ``pride'' in some contexts \\
• empathy → ``connection'', often ``compassion'', and whatever target emotion is empathized with (e.g., ``sadness'' for ``empathy for their loss'') \\
• disbelief → ``disgust'' if intense and explicitly directed toward misbehaving/failing third party \\
• awe → ``appreciation'' (could also point to ``fear'' in some contexts) \\
• curiosity or skepticism → *nothing* (these are just cognitive states, not emotions) \\
• concern for others or duty to care for others → ``compassion'' \\
\hspace*{1.5em} • compassion alone does not imply connection \\
• appreciating someone else (who is proud) does not imply that oneself is proud \\
• excitement → ``hope'', ``joy''
\normalsize
\end{tcolorbox}

\subsubsection{\colorbox{aesthetic-color}{Aesthetic Feeling}}
\begin{tcolorbox}[colback=brown!5!white,colframe=brown!50!black,title=Taxonomy Classification Annotation Guide,breakable]
\small
• empathy, skepticism, concern, admiration, frustration, relatability, exasperation, satisfaction, reassurance, validation, disappointment, compassion, discomfort → ``other'' if there is not a strong signal for one of our provided labels (these are other kinds of feelings not covered by our set of aesthetic feeling labels) \\
• interest in finding connections between events / piecing them together → ``curiosity'' \\
• if something grabs or holds focus → ``attention\_engagement'' \\
• nostalgia, vivid → ``evocation'' \\
• tension, anxiety, fear about future event → ``suspense'' \\
• being pulled into, drawn into, absorbed, immersed in a story; visceral, secondhand feelings (e.g. secondhand embarrassment) → ``transportation'' \\
• if not accompanied by another label, put ** (we don’t count empathy as an aesthetic feeling in and of itself). \\
• if you’ve already found one label, don’t feel the need to put ‘other’ to cover other aspects of the same response
\normalsize
\end{tcolorbox}

\subsubsection{\colorbox{prediction-color}{Prediction}}
\begin{tcolorbox}[colback=brown!5!white,colframe=brown!50!black,title=Taxonomy Classification Annotation Guide,breakable]
\small
• subject identification \\
\hspace*{1.5em} • prediction about narrator → ``narr\_*'' \\
\hspace*{1.5em} • prediction about character besides narrator → ``other\_char\_*'' \\
\hspace*{1.5em} • prediction about non-character entity → ``non\_char\_thing\_*'' \\
• action/event vs. state \\
\hspace*{1.5em} • active behavior / doing something → action/event \\
\hspace*{1.5em} • passive condition / being in a situation / feeling a certain way → state \\
\hspace*{1.5em} • there is a spectrum between states and actions, with many predictions falling somewhere in the middle. When in doubt, apply both labels (e.g., ``narr\_future\_state'' and ``narr\_future\_action'') \\
\hspace*{1.5em} • if the prediction is about someone ``continuing'', your label should be determined by the expected length of the continuation \\
\hspace*{3em} • continuing to argue/justify/make a point in this specific conversation → action \\
\hspace*{3em} • continuing a behavior indefinitely → action and state \\
\hspace*{3em} • continuing to feel a certain way or maintain a condition → state \\
• focus on the main point — if the prediction is about a future action, say action (even if some state must implicitly motivate that action) \\
• if a character is described as passively receiving something / something happening to them, focus on the action of the other character offering / doing something. \\
\normalsize
\end{tcolorbox}

\subsubsection{\colorbox{causal-color}{Causal Explanation}}
\begin{tcolorbox}[colback=brown!5!white,colframe=brown!50!black,title=Taxonomy Classification Annotation Guide,breakable]
\small
distinguishing characters from things: \\
• *character* (narr or other\_char): intentional/conscious actions, emotional reactions, behaviors, and mental states are associated with characters. \\
\hspace*{1.5em} • in addition to individuals, any group (e.g., family), animal, or company whose agency is foregrounded counts as a characters \\
• *thing*: cultural/institutional force (e.g., religious doctrine), systems, non-conscious processes, non-conscious body parts, objects. \\
\hspace*{1.5em} • underlying somatic or environmental factors, such as an undiagnosed medical conditon or instincts, that influence characters’ mental states or actions \\
\hspace*{1.5em} • cultural artifacts (e.g., books, films, video games): treat are considered things UNLESS creator agency is explicitly foregrounded \\
• special case: if a character’s behavior is described as enacting or being influenced by a social norm or cultural institution (e.g. religious doctrine), we consider that BOTH a character's action and a thing. \\

distinguishing narrator from either other character or things: \\
• narr: Text explicitly presents narrator's reasoning/beliefs/mental processes \\
• other\_char\_or\_thing: Reasoning attributed to non-narrator character or systemic factors \\
• KEY: Don't use narrator labels when narrator simply reports facts about others \\

general tips: \\
• check for multiple explanation types. Use multiple labels when more than one character or thing is being explained, or when multiple characters/things are doing the explaining. \\
• if someone or something is explained (partially or fully) by a character’s belief, perception, or opinion, use: \\
\hspace*{1.5em} • ``*\_explained\_by\_narr'' if the narrator holds the belief \\
\hspace*{1.5em} • ``*\_explained\_by\_other\_char\_or\_thing'' if another character holds the belief \\
\hspace*{1.5em} • if the belief explicitly stems from a cultural or institutional source (e.g., religious doctrine) or social norm, also include ``*\_explained\_by\_other\_char\_or\_thing'' \\
\hspace*{1.5em} • if the belief is a proposition (e.g., ``narrator believes President is angry''), label based on the belief’s content as well, if that content isn’t already captured. For example: ``\_explained\_by\_narr'' + ``\_explained\_by\_other\_char\_or\_thing'' \\
• when the narrator makes a neutral observation, comment, or mention, label based on what is described, not the act of commenting UNLESS the explanation itself is about emphasizes why/how the narrator made the comment or took the action. \\
• if the narrator emphasizes, argues, or believes something, include labels for both the action (e.g., arguing) and the content or subject of that action. \\
• other\_char\_or\_things (e.g., other char actions, non-conscious bodily processes in the narrator like illnesses, aesthetic qualities of comments) attributed (at least partly) to conscious actions, perceived beliefs, or expected reactions of the narrator → ``other\_char\_or\_thing\_explained\_by\_narr'' \\
• edge case: If it's unclear whether to label based on the character experiencing something or the one causing it, label the more active party. Focus on explaining their behavior, not the recipient’s experience. \\
• edge case: If it's ambiguous whether a character is the narrator or another character, assume they are not the narrator. Label as another character.
\normalsize
\end{tcolorbox}

\subsubsection{\colorbox{moral-color}{Moral}}
\begin{tcolorbox}[colback=brown!5!white,colframe=brown!50!black,title=Taxonomy Classification Annotation Guide,breakable]
\small
general tips: \\
• select labels based on thematic relevance, even if the text is not fully endorsing the value (e.g., if the text highlights a tradeoff between multiple values) \\

label-specific tips: \\
• independent thinking, critical evaluation to form one's own opinion, effortful researching to make informed personal choices, creative problem-solving \\
• embracing challenges and change FOR THEIR OWN SAKE, risk-taking, excitement and adventure, unpredictability/adaptability, valuing novelty and difficulty as inherently rewarding → ``stimulation'' \\
• pleasure, enjoyment, fun, humor, entertainment, appearance → ``hedonism'' \\
• demonstrated individual competence, measurable goal attainment and success, skill development leading to improved performance, professional or competitive success with clear outcomes (focus on results and competence, not just effort or persistence) → ``achievement'' \\
• authority, institutional influence/control, status and dominance dynamics, organizational hierarchy issues, obedience, submission → ``power'' \\
• safety and risk management for SIGNIFICANT threats, relationship survival, safeguarding financial or health status from substantial harm, maintaining equanimity against serious risks → ``security'' \\
• following communication and social interaction norms, anticipating and trying to avoid misunderstandings → ``conformity'' \\
• respecting the specific customs and practices according to cultural consensus or a large religious or state institution, respecting or learning from the past, trusting conventional media/institutions (e.g. news, libraries) → ``tradition'' \\
• caring for family, friends, teammates; loyalty and commitment to specific groups, helping those in close proximity → ``benevolence'' \\
• fairness and equality, appreciating/celebrating differences or variation across individuals or groups, avoiding discrimination or prejudice based on assumptions about unknown others, broad social welfare concerns → ``universalism'' \\

edge cases: \\
• if a text highlights awareness of variation across traditions, label ``universalism'', not ``tradition''—which should be used when a text focuses on a single tradition (either positively or critically). \\
• referring to systemic things or to the fact that individuals are explained by social factors does not automatically imply ``universalism'' \\
• posts about pleasure vs. practicality (e.g., choosing practical/reliable option over flashier option) can still merit ``hedonism'', even if pleasure-seeking isn’t explicitly endorsed. \\
• posts about pleasure vs. practicality (e.g., choosing practical/reliable option over flashier option) can still merit ``hedonism'', even if pleasure-seeking isn’t explicitly endorsed. \\
• minor miscommunications, everyday foibles, or small issues with tools/resources do not warrant a ``security'' label unless there’s a clear threat to well-being. \\
• critiques/descriptions of large institutions: \\
\hspace*{1.5em} • use ``power'' to reflect dynamics of control or abuse.
\hspace*{1.5em} • add ``security'' if societal risks are emphasized.
\hspace*{1.5em} • add ``achievement'' if poor competence, mismanagement, or flawed strategy is central.
• strategy, planning, and persistence are usually ``achievement'' (if goal-oriented) or ``self-direction'' (if analytical), NOT ``stimulation'' unless the challenge itself is valued as rewarding \\
• research and verification activities are ``self-direction'' when about critical thinking; ``universalism'' when about avoiding assumptions about unknown others \\
• KEY: if in doubt about whether to include a label that is not clearly implied or covered by the definitions or guidelines, leave it out. 
\normalsize
\end{tcolorbox}

\subsubsection{\colorbox{stance-color}{Stance}}
\begin{tcolorbox}[colback=brown!5!white,colframe=brown!50!black,title=Taxonomy Classification Annotation Guide,breakable]
\small
answer based on the keywords in the beginning of the statement: \\
• support → ``support\_belief\_norm'' \\
• counter → ``counter\_belief\_norm'' \\
• be neutral to → ``neutral\_belief\_norm''
\normalsize
\end{tcolorbox}

\subsubsection{\colorbox{character-color}{Character Appraisal}}
\begin{tcolorbox}[colback=brown!5!white,colframe=brown!50!black,title=Taxonomy Classification Annotation Guide,breakable]
\small
determine the sentiment of appraisal based on the keywords at the beginning of the statement: \\
• positively → ``positive\_appraisal\_*'' \\
• negatively → ``negative\_appraisal\_*'' \\
• neutrally → ``neutral\_appraisal\_*'' \\

general: \\
• if it is ambiguous whether the character being judged is the narrator or another character, assume it is another character
\normalsize
\end{tcolorbox}

\subsection{Inference Classification Validation}
\label{app:tax_class_validation}

\subsubsection{Inter-Annotator Agreement}
\begin{figure*}[t]
    \centering
    \includegraphics[width=\linewidth]{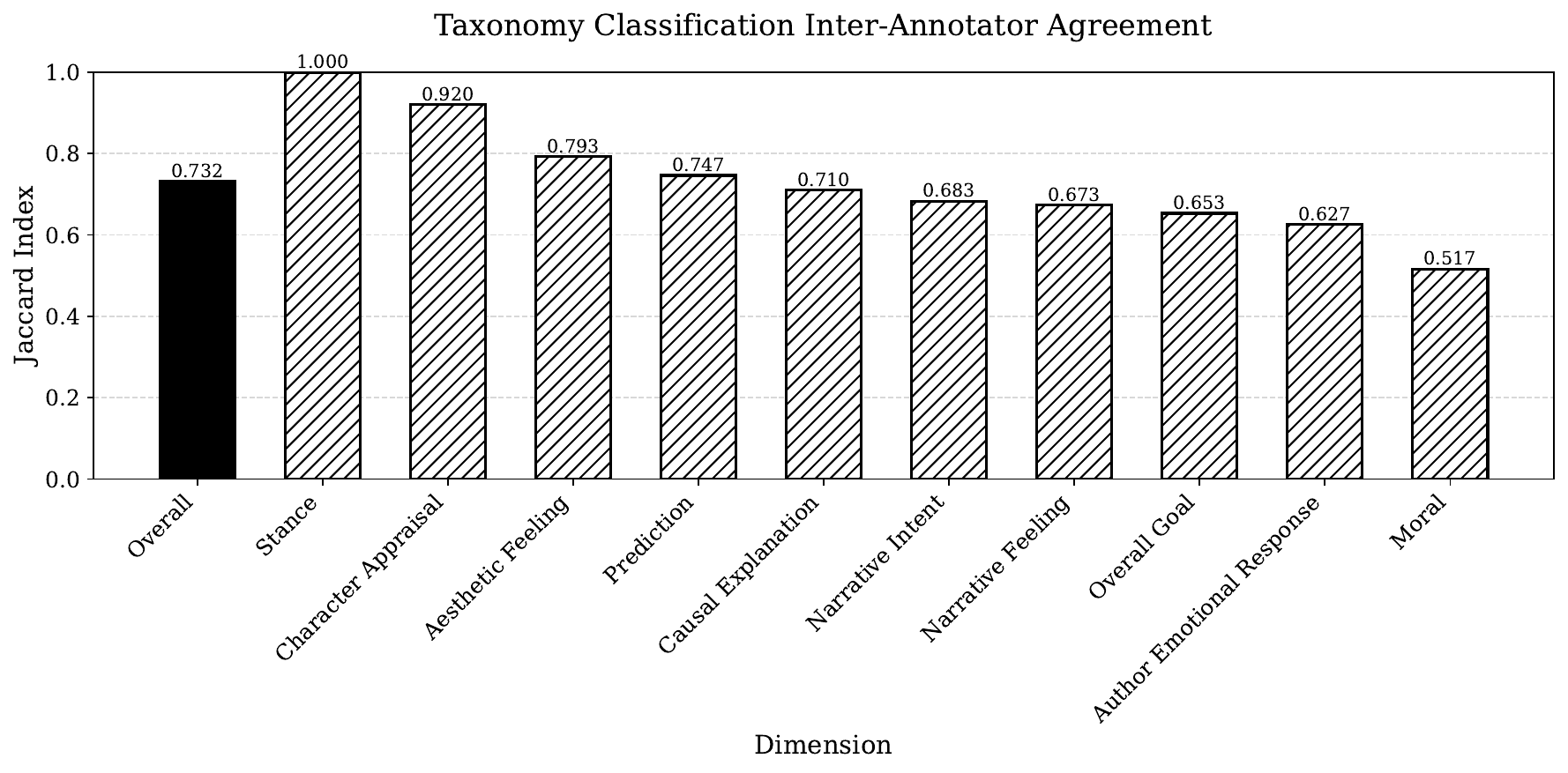}
    \caption{Inter-rater Agreement using Jaccard Index for Taxonomy Classification. Values represent average Jaccard indices across 50 instances per dimension.}
    \label{fig:tax-class-iaa}
\end{figure*}

Fig. \ref{fig:tax-class-iaa} plots the Jaccard Indices between the two annotators for each dimension.

\subsubsection{Annotation Protocol Details}
\label{app:tax-class-protocol}
The first author annotated $N=100$ examples per taxonomy dimension to refine guidelines and k-shot sampling strategy, validating with leave-one-out cross-validation. After reaching Macro F1 scores above $0.8$ on all dimensions, the finalized guidelines were used to annotate test instances. To address labeling skew and the large number of subdimensions, we included only those that appeared at least twice in the $100$ validation examples and at least once in the first $100$ test examples. If a subdimension had fewer than $5$ test labels, annotation continued until each had at least $5$.

\subsubsection{Error Analysis}
We qualitatively inspect incorrect multi-label inference classification predictions from \ssfGenerator to identify opportunities for refinement in future work. We find that many error cases reflect fundamental ambiguities that arise in the act of categorization along gradients, e.g., distinguishing states from actions. Thus, some failures highlight the intrinsic challenges of defining discrete categories for dimensions of reader response, rather than easily correctable shortcomings of the classifier itself.

\paragraph{\textsc{overall\_goal}}
\ssfClassifier sometimes confuses informational support with persuasion, treating statements as solely informative even when a subtle persuasion element, such as an innocuous implicit recommendation, is present. It can also struggle to recognize certain implicit requests for emotional connection, e.g., when a story is told in part to ``connect'' with others.

\paragraph{\textsc{narrative\_intent}} There are no stark trends, but we observe some parallels with the errors for the overall goal dimension. For example, we observe failures to recognize demonstrations of emotional support needs, such as when a strong negative emotion is expressed. Moreover, the model is overly reliant on the presence of the term ``personal'' in predicting the show\_identity sublabel, which is not always applicable.

\paragraph{\textsc{author\_emotional\_response}} A common failure traces back to \ssfClassifier seemingly failing to model our expansive definition of the ``relief'' sublabel, which could apply in cases like clarifying a misunderstanding.

\paragraph{\textsc{causal\_explanation}}
\ssfClassifier struggled when it was unclear whether a cause should be attributed to a person’s internal beliefs or to an external world state—for example, when behavior stems from a particular interpretation of the world. Such distinctions are often subtle, so expert annotators expected the model to have difficulty with these cases.

\paragraph{\textsc{prediction}} These errors often trace back to challenges in distinguishing between actions and states, a binary distinction encoded in our discrete prediction sublabels. Human annotators similarly struggled to draw a boundary between actions and states. How to enable a categorical analysis that parsimoniously delineates the gradient between state and action remains an open challenge.

\paragraph{\textsc{moral}} The primary failure pattern was overprediction (false positives). Additionally, predictions seem cued by relatively surface-level keywords rather than underlying values, perhaps a consequence of the gap between colloquial meanings of the sub-labels and their more abstract and broad meanings as theoretical constructs in \citeposs{Schwartz2012-xk} theory of values.

\paragraph{\textsc{narrative\_feeling}}
The major failure pattern was a tendency for \ssfClassifier to predict too many categories (false positives). This suggests a systemic bias in the average number of labels per inference, rather than a bias toward particular labels.

\paragraph{\textsc{aesthetic\_feeling}} We observe no clear trends. However, we observe that \ssfClassifier fails to recognize certain phrases, e.g. ``drawn into'' as keywords indicative of the narrative transportation sublabel. 

\section{\ssfSimMetric Details}
\label{app:ssf-sim-details}
To compare storytelling communities through the lens of \ssfFormalisms, we combine the reader-response predictions from \ssfGenerator and \ssfClassifier into a similarity metric, \ssfSimMetric. For each taxonomy dimension, \ssfSimMetric computes (1) the cosine similarity between subreddit-level \textit{inference embeddings},\footnote{The structured natural language inferences on which \ssfSimMetric is based reflect the 10 dimensions of reader response in \ssfTaxonomy. Although we use Sentence-BERT to measure similarity as part of \ssfSimMetric, the input texts are the generated inferences, not the stories. Thus, the notion of semantic similarity from Sentence-BERT that partially comprises \ssfSimMetric predominately reflects similarity at the level of reader response, instead of explicit story content.} and (2) the Jensen–Shannon–based similarity between categorical \textit{sublabel distributions}. We derive subreddit-level representations by mean-pooling inference embeddings and normalizing sublabel counts into probability distributions. The resulting dimension-level similarities are aggregated into a composite metric that enables abstract comparison of narrative practices across communities.

For the categorical component (\ssfClassifier outputs), we convert subdimension counts into probability distributions and compute the complement of the Jensen–Shannon distance to obtain dimension-level similarities, averaging across all dimensions for an aggregate score.

For the generation-based component  (\ssfGenerator outputs), we represent each subreddit by mean-pooling the embeddings of the inference template variables at the subreddit level. Cosine similarities are then computed between the corresponding embeddings for each subreddit pair. We average these similarity scores first within each dimension (across template variables) and then across dimensions to obtain a second aggregate similarity score. 

Finally, we combine the two components into a composite relative similarity measure by taking a weighted sum of each subreddit pair’s ranks across submetrics.\footnote{In our analysis, we upweight the categorical component (0.667) to emphasize similarity based on the abstract \ssfTaxonomy sublabels, relative to the structured inferences, which encode a degree of semantic information.}

\subsection{Formal Definition}
Let $t_{i,j,k}$ be the text sequence value for the $k^{th}$ variable slot of the $j^{th}$ dimension template for the $i^{th}$ story. 

Let $\phi$ be an embedding model that maps text sequences to dense vector representations. We compute $L_2$-normalized embeddings for all $i,j,k$: \[
h_{i,j,k} = \frac{\phi(t_{i,j,k})}{\|\phi(t_{i,j,k})\|_2}
\].

Through mean-pooling, we compute a single embedding for each dimension, $h_{i,j}$:
\begin{align}
h_{i,j} &= \frac{1}{K_j}\sum_k h_{i, j, k}
\end{align}
where $K_j$ is the number of slots in the template for dimension $j$.

Next, we again use mean-pooling to compute a community-level representation of a taxonomy dimension, $h_{c,j}$:
\begin{align}
h_{c,j} &= \frac{1}{I_c}\sum_i h_{i, j} \\
\end{align}

To compare two communities, we compute the average cosine similarity between taxonomy dimensions:
\begin{align}
\texttt{ssf-sim}_{gen}(c, c') 
    &= \frac{1}{10} \sum_j 
       \cos(h_{c,j}, \, h_{c',j})
\end{align}
where $10$ corresponds to the number of dimensions in \ssfTaxonomy.

For the categorical component of the metric, let $S_j$ be the number of subcategories in taxonomy dimension $j$, and let $\text{count}_{c,j,s}$ denote the number of stories in community $c$ assigned to subcategory $s$ of dimension $j$. We define the normalized probability distribution over subcategories for community $c$ as:

\begin{align}
p_{c,j,s} &= \frac{\text{count}_{c,j,s}}{\sum_{s'=1}^{S_j} \text{count}_{c,j,s'}} 
\quad \text{for } s = 1, \dots, S_j
\end{align}

The full distribution over subcategories for dimension $j$ is then

\begin{align}
p_{c,j} = \{ p_{c,j,1}, \dots, p_{c,j,S_j} \}.
\end{align}

Using these distributions, the categorical similarity between two communities $c$ and $c'$ is computed as the complement of the Jensen-Shannon divergence \cite{Lin1991-zq} averaged over all dimensions:

\begin{align}
\small
\texttt{ssf-sim}_{class}(c, c') 
    &= \frac{1}{10} \sum_{j}
       ( 1 - JSD(p_{c,j}, \, p_{c',j}))
\end{align}

We define rankings of community pairs by similarity, $\pi_{\texttt{ssf-sim}_{gen}}$ and $\pi_{\texttt{ssf-sim}_{class}}$, and then integrate these into a composite ranking:

\begin{align}
\texttt{ssf-sim}(c, c') &= \lambda\pi_{\texttt{ssf-sim}_{class}}(c, c') \\
&+ (1 - \lambda)\pi_{\texttt{ssf-sim}_{gen}}(c, c')
\end{align}
where $\lambda$ interpolates the importance of the sub-metrics.

We combine ranks rather than raw scores because the metrics have different ranges: [-1, 1] for cosine similarities and [0, 1] for JSD. We interpret the composite \texttt{ssf-sim} scores \textit{relatively} in the context of a ranking, $\pi_{\texttt{ssf-sim}}$.

\subsection{Global Validation}
\label{app:ssf-sim-global-validation}
\ssfSimMetric is intended to facilitate studying the social functions of storytelling through the lens of reader response, beyond explicit story content. Accordingly, our approach to global validation balanced two goals: encouraging annotators to consider similarity at pragmatic and interpretive levels, while avoiding undue bias toward \ssfSimMetric’s specific modeling choices. To this end, annotators judged similarity using a small set of abstract, content-agnostic categories (author intent and affective responses, causes/effects, reader affective responses, and readers’ normative judgments) rather than the full 10 \ssfTaxonomy dimensions that \ssfSimMetric directly models.

Concretely, two annotators evaluated a random sample of pairs of pairs of stories to determine which pair was more similar according to the following annotation guide:

\begin{tcolorbox}[
  colback=gray!5!white,
  colframe=gray!75!black,
  title=Story Similarity Annotation Guide,
  breakable
]
\small
Decide which pair of stories is more similar according to the following abstract criteria: \\
• author intent(s) \\
• author affective responses \\
• readers’ affective responses \\
• causes/effects (e.g., of character behavior, events) \\
• readers’ normative judgments about characters
\vspace{1.5em}

Do not make comparisons based solely on explicit content overlap in the story. Instead, approach these comparison categories abstractly in terms of the communicative functions, emotional impacts, archetypal character goals/drives and events, and basic judgments (e.g., approval/disapproval) about characters.
\end{tcolorbox}
\normalsize

\begin{figure}[h]
    \centering
    \includegraphics[width=\linewidth]{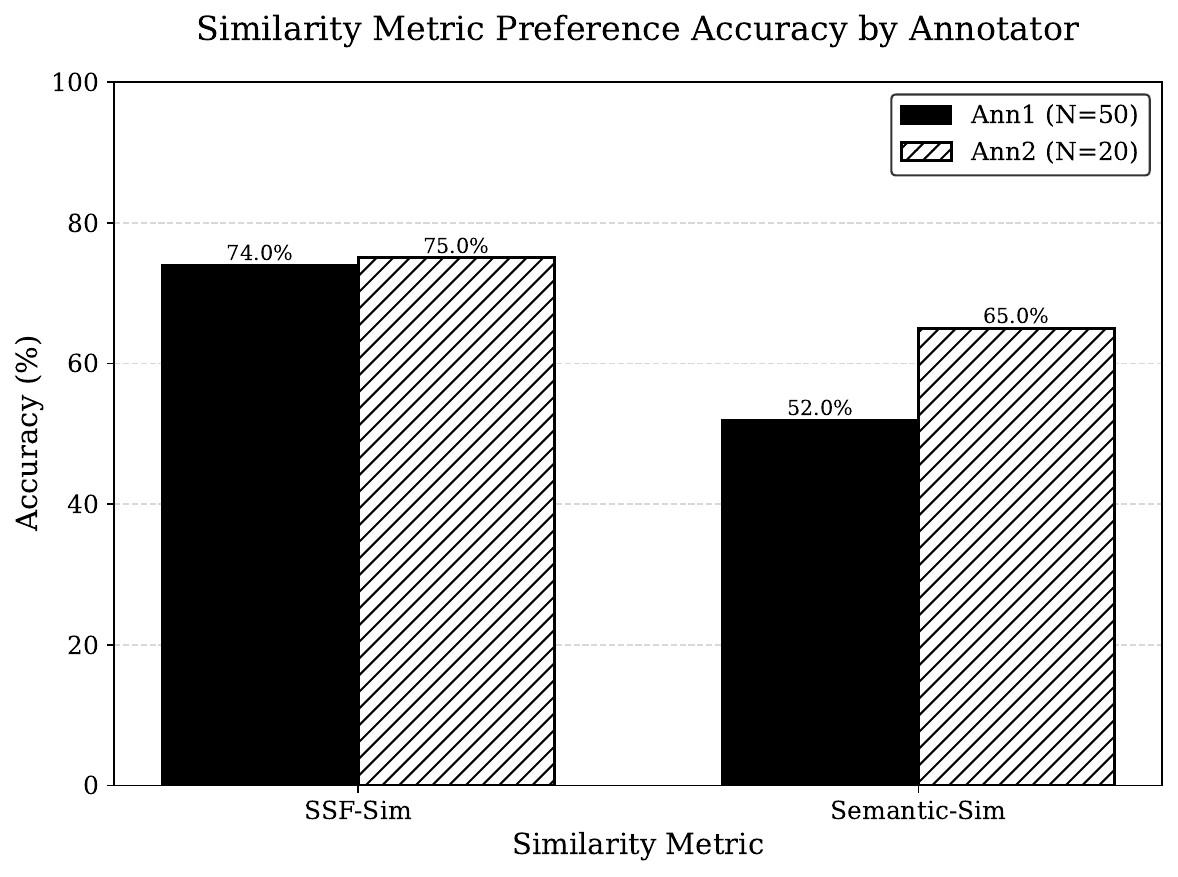}
    \caption{How well \ssfSimMetric and standard semantic similarity metrics recover human preferences for story similarity.}
    \label{fig:ssf-sim-global-val}
\end{figure}

The results from the small-scale annotation are reported in Fig. \ref{fig:ssf-sim-global-val}.

\subsection{Examples: \ssfSimMetric versus Semantic Similarity}
\label{app:ssf-sim-vs-sem-examples}
Table \ref{tab:story_pair_quadrants} provides example story pairs demonstrating how \ssfSimMetric relates to semantic similarity measures.

\onecolumn
\small
\begin{longtable}{p{3.5cm} p{5.5cm} p{5.5cm}}
\caption{
    Paraphrased story pairs sampled from the quadrants in Fig. \ref{fig:similarity}, using $1,000$ random cross-subreddit story pairs ranked by \ssfSimMetric and Sentence-BERT \texttt{all-MiniLM-L6-v2} based semantic similarity. 
    \textbf{ssf-sim $\uparrow$ / semantic-sim $\downarrow$}: Pairs where the \ssfSimMetric rank is much higher than the semantic similarity rank, indicating similar intent/reception but different topics.
   \textbf{ssf-sim $\downarrow$ / semantic-sim $\uparrow$}: Pairs where the semantic similarity rank is much higher than \ssfSimMetric, indicating different narrative intent/reception but similar topics.
    \textbf{ssf-sim $\uparrow$ / semantic-sim $\uparrow$}: Pairs with high similarity on both dimensions (low average rank).
    \textbf{ssf-sim $\downarrow$ / semantic-sim $\downarrow$}: Pairs with low similarity on both dimensions (high average rank).
}
\label{tab:story_pair_quadrants}\\
\toprule
\textbf{Subreddit Pair} & \textbf{Story 1} & \textbf{Story 2} \\
\midrule
\endfirsthead
\multicolumn{3}{c}{\tablename\ \thetable\ -- continued from previous page} \\
\toprule
\textbf{Subreddit Pair} & \textbf{Story 1} & \textbf{Story 2} \\
\midrule
\endhead
\midrule
\multicolumn{3}{r}{Continued on next page} \\
\endfoot
\bottomrule
\endlastfoot
\midrule
\multicolumn{3}{l}{\textbf{\ssfSimMetric $\uparrow$ / semantic-sim $\downarrow$}}\\
\midrule
\texttt{r/MakeupAddiction} - \texttt{r/buildapc} & i might be wrong but i think i might have had one that was made by Ulta. there is also one called lip shimmer mocha from designer skin it was a click up pencil with brush tip! heres a walmart link i found to a listing for one [walmart link](<url>) & i got mine to <number> ghz with evga clc360 in a h500i case and for max temps of \textbackslash{}textasciitilde{}<number>-70c so i took it down to <number> ghz and now max is like 50c. forgot what voltage is but it's pretty low edit: just checked my voltage and it's maximum is <number>.20v with a min of <number>.60v \\
\midrule
\multicolumn{3}{l}{\textbf{\ssfSimMetric $\downarrow$ / semantic-sim $\uparrow$}}\\
\midrule
\texttt{r/funny} - \texttt{r/news} & doesn't seem so, he does appear to do something with his hand or his leg in the end, either a strained leg or he's just gesticulating to the driver to how close he was to backing into the other car. & no, the cops claimed that it was reported as a robbery. **there wasn't a robbery** the cops lied about there being one as an excuse for pulling the womans black grandson out of the car at gunpoint. They were so racist to believe that a black teen in the car with old white women means "that's a car jacking." \\
\midrule
\multicolumn{3}{l}{\textbf{\ssfSimMetric $\uparrow$ \& semantic-sim $\uparrow$}}\\
\midrule
\texttt{r/CFB} - \texttt{r/nfl} & more than anything i put this on the refs for allowing it. if you watch, while the ball is moving, he is waving his hands at his waist, which apparently he asked the ref about before the play. that wouldn't be a new rule though, it would just be enforcing the current rule. & he grabbed his crotch after scoring a touchdown on the same drive a guy launched at his head and smacked him in the face after the whistle and all he did was make a gesture at the sidelines, they're not the same at all \\
\midrule
\multicolumn{3}{l}{\textbf{\ssfSimMetric $\downarrow$ \& semantic-sim $\downarrow$}}\\
\midrule
\texttt{r/buildapc} - \texttt{r/politics} & so i just finished my first ever build. Everything is working fine but i get no signal. i'm trying to use my tv until my monitor arrives. do i need a monitor for the initial boot? & broaddrick's voice was heard under oath. why won't conservatives give ford the same justice from wikipedia: in a sworn statement in <number> with the placeholder name "jane doe \#<number>",[<number>] broaddrick filed an affidavit with paula jones' lawyers saying there were unfounded rumors and narratives circulating "that mr. clinton had made unwelcome sexual advances toward me in the late seventies... these allegations are untrue".[<number>] \\
\end{longtable}

\twocolumn
\normalsize

\subsection{Comparison to Alternative Narrative Similarity Measures}
Below, we situate \ssfSimMetric in relation to several recently proposed narrative similarity metrics.

\citet{Shen2023-pj} model empathic similarity across three dimensions: main events, emotional reactions, and morals. Like \ssfSimMetric, their model-based metric attends to extra-textual perceptions that extend beyond semantic similarity. However, its scope is considerably narrower, overlapping with just $2/10$ \ssfTaxonomy dimensions.

\citet{Johnson2025-kg} derive a similarity measure from author-provided metadata on relatively long-form fanfiction stories. Their approach shares \ssfSimMetric’s interest in moving beyond surface-level signals but depends on explicit annotations. As a result, it cannot be applied to datasets like the \ssfCorpus, which lack such metadata. In contrast, \ssfSimMetric centers reader response, is designed for short-form stories in online communities, and produces rich, structured inferences without requiring metadata, making it broadly applicable across social media and other metadata-sparse storytelling domains.

\citet{Hatzel2024-cg} use contrastive learning over multiple retellings or summaries of the same story. Their notion of similarity is likely to emphasize plot-level overlap, in contrast to \ssfSimMetric’s focus on readers’ inferences and perceptions.

\section{Additional Results}
\label{appendix-additional-results}

\subsection{\ssfTaxonomy Sublabel Frequency Distributions}
\begin{figure*}[t]
    \centering
    \includegraphics[width=\linewidth]{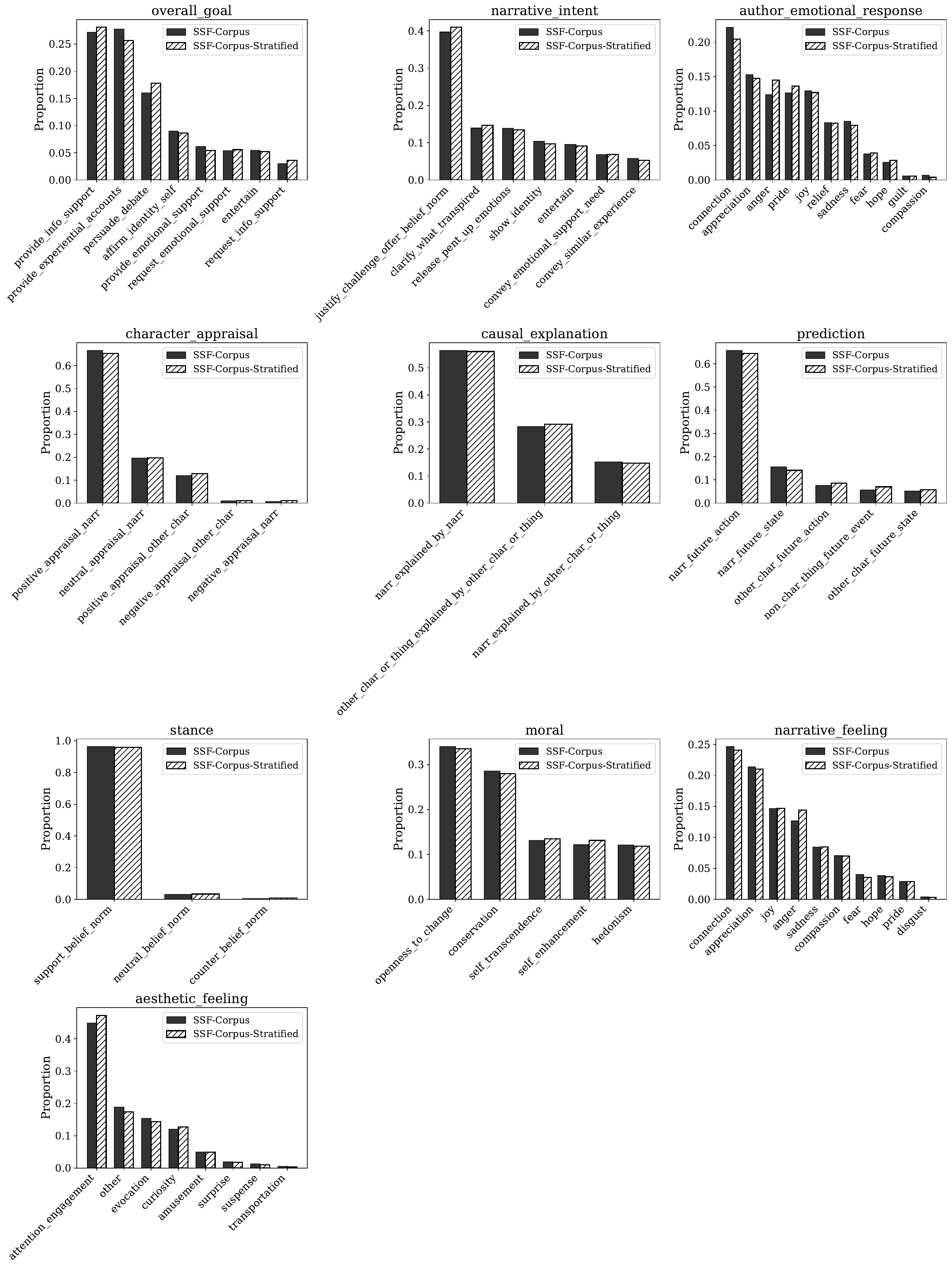}
    \caption{Bar plots showing the \ssfTaxonomy dimension-level sublabel distributions for \ssfCorpus and \ssfStratifiedCorpus.}
    \label{fig:all_freqs}
\end{figure*}

See Fig. \ref{fig:all_freqs}.

\subsection{Associations between overall goals and narrative intents}
\begin{figure}[t]
    \centering
    \includegraphics[width=\columnwidth]{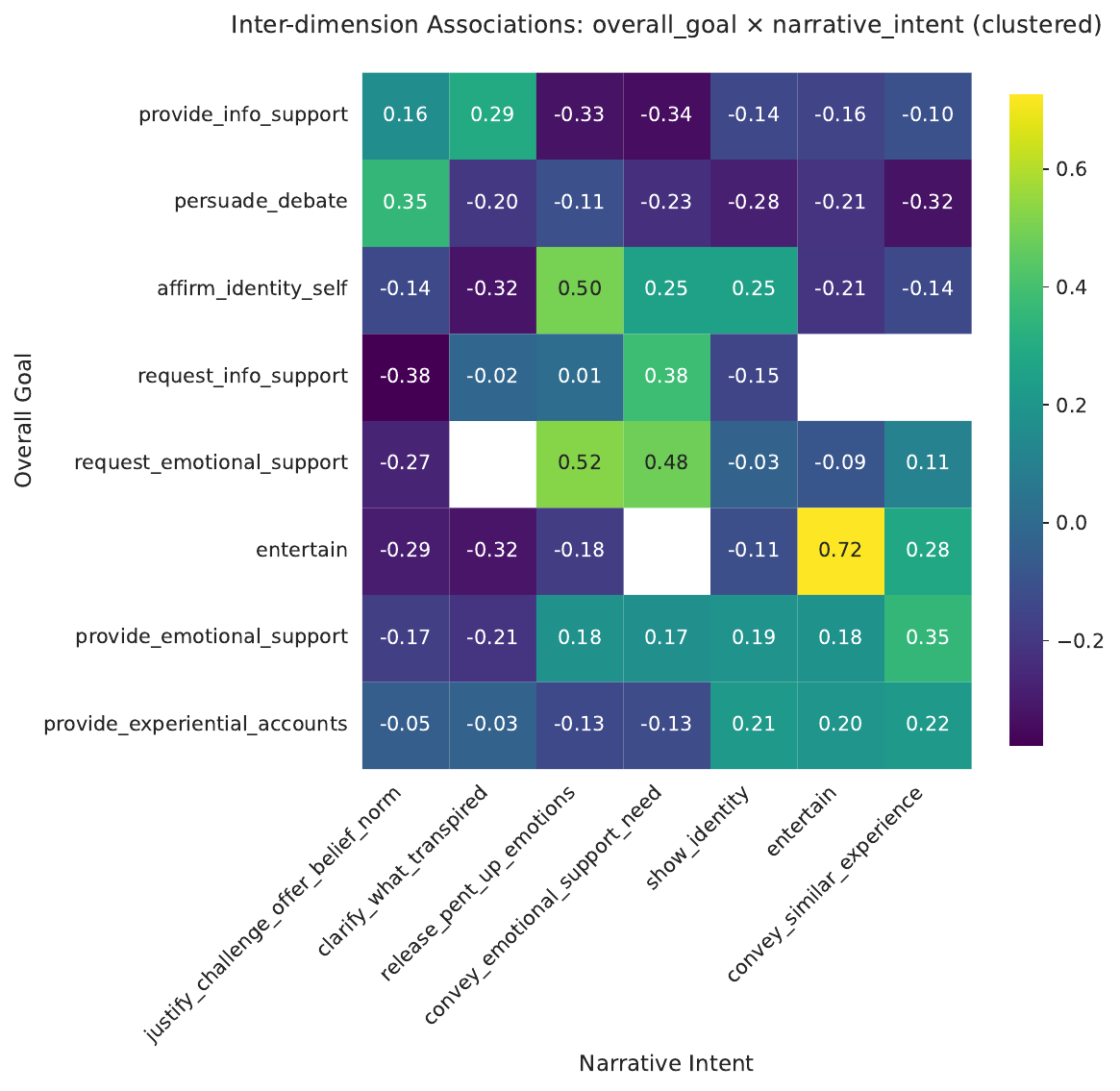}
    \caption{Normalized pointwise mutual information (NPMI) between overall goals and narrative intents.}
    \label{fig:npmi}
\end{figure}

See Fig. \ref{fig:npmi}.

\subsection{Example comparison of subreddits with high vs. low narrative diversity along author-centric dimensions}
\begin{figure*}[t]
    \centering
    \includegraphics[width=\linewidth]{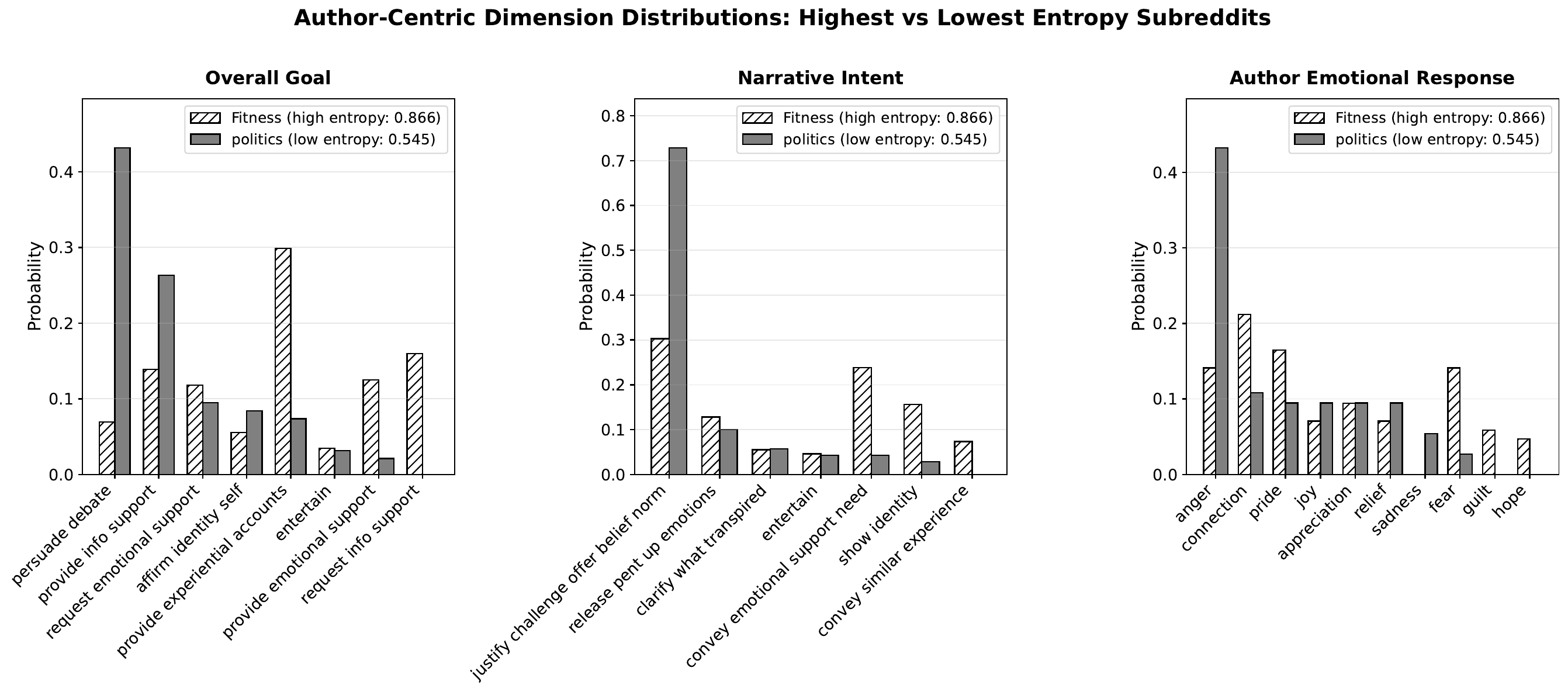}
    \caption{Comparison between subreddits with the highest (\texttt{r/Fitness}) vs. lowest (\texttt{r/politics}) narrative diversity along the author-centric dimensions of \ssfTaxonomy.}
    \label{fig:auth_entropy}
\end{figure*}

See Fig. \ref{fig:auth_entropy}.

\section{Responsible NLP Checklist}
\paragraph{Artifact Use Consistent With Intended Use} 
\ssfGenerator and \ssfClassifier are built on Llama 3.1 base models. Distribution of our finetuned models is therefore bound by the Llama 3.1 license.\footnote{\url{https://www.llama.com/llama3_1/license/}}

\paragraph{Model Budget}
Approximately 100 NVIDIA RTX A6000 GPU hours are required to replicate our results. 

\pagebreak

\end{document}